Review

# Unsupervised anomaly detection in time-series: An extensive evaluation and analysis of state-of-the-art methods

Nesryne Mejri [a,*], Laura Lopez-Fuentes [a], Kankana Roy [a], Pavel Chernakov [a], Enjie Ghorbel [b,a], Djamila Aouada [a]

[a] *Interdisciplinary Centre for Security, Reliability and Trust (SnT), University of Luxembourg, 6 Rue Richard Coudenhove-Kalergi, Luxembourg-City, L-1359, Luxembourg*
[b] *Cristal Laboratory, National School of Computer Science, University of Manouba, Campus Universitaire de La Manouba, Manouba, 2010, Tunisia*

A B S T R A C T

Unsupervised anomaly detection in time-series has been extensively investigated in the literature. Notwithstanding the relevance of this topic in numerous application fields, a comprehensive and extensive evaluation of recent state-of-the-art techniques taking into account real-world constraints is still needed. Some efforts have been made to compare existing unsupervised time-series anomaly detection methods rigorously. However, only standard performance metrics, namely precision, recall, and F1-score are usually considered. Essential aspects for assessing their practical relevance are therefore neglected. This paper proposes an in-depth evaluation study of recent unsupervised anomaly detection techniques in time-series. Instead of relying solely on standard performance metrics, additional yet informative metrics and protocols are taken into account. In particular, (i) more elaborate performance metrics specifically tailored for time-series are used; (ii) the model size and the model stability are studied; (iii) an analysis of the tested approaches with respect to the anomaly type is provided; and (iv) a clear and unique protocol is followed for all experiments. Overall, this extensive analysis aims to assess the maturity of state-of-the-art time-series anomaly detection, give insights regarding their applicability under real-world setups and provide to the community a more complete evaluation protocol.

**Contents**

* Corresponding author.
  *E-mail addresses:* nesryne.mejri@uni.lu (N. Mejri), l.lopez@totia.es (L. Lopez-Fuentes), kankana.roy@ki.se (K. Roy), pavel.chernakov.001@student.uni.lu (P. Chernakov), enjie.ghorbel@isamm.uma.tn (E. Ghorbel), djamila.aouada@uni.lu (D. Aouada).

## 1. Introduction

A multivariate time-series corresponds to a temporally ordered set of variables. This mathematical representation has been used in countless domains such as finance, health, and biomechanics. Designing methods for automatically analyzing time-series (e.g. forecasting, classification, anomaly detection) has been widely investigated by researchers (Mahalakshmi, Sridevi, & Rajaram, 2016; Zhao, Lu, Chen, Liu, & Wu, 2017). A particular focus is given to anomaly detection in time-series (Blázquez-García, Conde, Mori, & Lozano, 2021; Wu & Keogh, 2021). In general, an anomaly or outlier can be defined as an observation or sample that does not follow an expected pattern. The popularity of anomaly detection in time-series is probably due to its interest in numerous industrial contexts. As an example, one can mention the detection of faulty sensors (Cook, Mısırlı, & Fan, 2019), fraudulent bank transactions (Devaki, Kathiresan, & Gunasekaran, 2014), and pathologies in medical data (Keogh, Lin, Fu, & Van Herle, 2006; Kourou, Exarchos, Exarchos, Karamouzis, & Fotiadis, 2015).

In the literature, some attempts have been made to develop supervised and semi-supervised approaches (Golmohammadi & Zaiane, 2017; Jiang, Kao, & Li, 2021). Although supervised techniques may achieve higher detection performance on anomalies seen during training, they usually risk overfitting to those anomalies, resulting in poor generalization to novel outliers. Semi-supervised approaches offer a more flexible solution leveraging both labeled and unlabeled data (Oliver, Odena, Raffel, Cubuk, & Goodfellow, 2018). However, despite being promising (Oliver et al., 2018; Zhou et al., 2022), these methods still rely on a certain amount of annotated data, which can be constraining. Hence, the task of time-series anomaly detection is usually formulated as an unsupervised problem (Su et al., 2019). In fact, since anomalies occur rarely, annotating data becomes challenging and costly; This makes unsupervised learning more adequate despite being exposed to additional challenges such as the lack of explicit guidance and complex hyper-parameter tuning (Garg & Kalai, 2018). In this article, we focus on the topic of unsupervised anomaly detection in time-series.

Earlier methods of anomaly detection in time-series mostly employed traditional Machine Learning (ML) (Jin, Chen, Li, Poolla, & Sangiovanni-Vincentelli, 2019; Liu, Ting, & Zhou, 2008) and auto-regressive (Chen, Wang, Wei, Li, & Gao, 2019; Yaacob, Tan, Chien, & Tan, 2010) techniques. However, as discussed in Choi et al. (2021), these approaches are mainly subject to *the curse of dimensionality*. In other words, their performance drops in the presence of high-dimensional time-series.

To address this, motivated by the tremendous advances in Deep Learning (DL), massive efforts have been recently made to design suitable Deep Neural Network (DNN) architectures (Deng & Hooi, 2021; Su et al., 2019; Xu et al., 2018). These DL-based approaches have achieved impressive performance in terms of *standard performance metrics* (precision, recall, and F1-score). Nevertheless, despite their promising results, their suitability in a realistic industrial context still needs further investigation. For that purpose, it is timely to propose an extensive comparison of recent unsupervised DL techniques that consider the following aspects:

(**i**) **Model size and model stability:** Existing methods overlook the *model size* and the *model stability*, which are important indicators of the scalability and the performance stability. By a stable model, we mean a model which has stable performance under different training trials.

(**ii**) **Unified experimental protocol:** There is no clear experimental protocol for evaluating state-of-the-art methods. As a consequence, it can be noted that the reported experimental values vary considerably from one paper to another. For instance, as highlighted by Kim et al. (2022), a peculiar evaluation protocol called Point Adjustment (PA) introduced by Xu et al. (2018) is often used (Audibert, Michiardi, Guyard, Marti, & Zuluaga, 2020; Su et al., 2019), while it is ignored in other cases (Li et al., 2019; Malhotra et al., 2015).

(**iii**) **Performance metrics for time-series:** As discussed by Tatbul et al. (2018), the used standard performance metrics (precision, recall, and F1-score) might not be entirely adequate for evaluating time-series anomaly detectors. These metrics were initially designed for time-independent predictions and not for range-based ones. As an alternative, Tatbul et al. (2018) extended these metrics to time-series. However, it can be noted that current state-of-the-art methods do not consider these relatively novel evaluation criteria.

(**iv**) **Experimental analysis with respect to the anomaly type:** a detailed experimental evaluation with respect to the type of anomaly is missing in the state-of-the-art. Significant efforts have been dedicated to rigorously defining the different possible types of outliers in time-series (Choi et al., 2021; Lai et al., 2021). However, no detailed experimental analysis has been carried out in that direction.

(**v**) **Comparison against ML methods:** Similar to the works of Wu and Keogh (2021) and Audibert et al. (2022), we emphasize the importance of comparing traditional ML strategies to DL approaches. Recent studies (Darban et al., 2022; Liu et al., 2022, 2023) tend to focus on DL-based works, often overlooking ML techniques. Our findings are consistent with Wu and Keogh (2021) and Audibert et al. (2022), and indicate that ML methods remain relevant for the task unsupervised anomaly detection in time-series.

In the literature, some survey studies were proposed for unsupervised time-series anomaly detection (Liu et al., 2023; Zhong et al., 2023). They primarily focus on presenting recent approaches, their relevant applications and their challenges and limitations. Some other works (Paparrizos et al., 2022; Wu & Keogh, 2021) have conducted experiments to identify the flaws of current benchmark datasets and scoring functions, proposing new datasets and issuing recommendations for practitioners. While few other evaluation studies focused on experimentally comparing recent anomaly detection algorithms (Audibert et al., 2022; Belay et al., 2023; Choi et al., 2021; Darban et al., 2022; Kim et al., 2022; Lai et al., 2021; Liu et al., 2022; Paparrizos et al., 2022; Schmidl et al., 2022). Our work belongs to this latest category. For instance, Choi et al. (2021) present a brief comparison of recent DL algorithms in terms of precision, recall, and F1-score but neglect the model size and model stability. We can also mention the work of Lai et al. (2021), where a new taxonomy for time-series outliers is proposed. Then, based on that, a methodology to generate synthetic datasets is suggested. They finally compare nine different algorithms according to outlier types but they do not include the latest DL algorithms. Nevertheless, similar to Choi et al. (2021), they only focus on classical evaluation criteria, omitting range-based evaluation, model stability and sizes. Furthermore, Kim et al. (2022) present a rigorous evaluation of recent DL techniques by questioning the Point Adjustment protocol. Nevertheless, the model size and model stability, as well as the performance metrics for time-series are not considered. Schmidl et al. (2022) propose a large-scale evaluation study of existing anomaly detection methods, thereby assessing the overall progress made in this field. Nevertheless, they do not investigate the conceptual differences and limitations of different types of approaches. In addition, recent state-of-the-art deep learning methods published in top-tier venues such as Audibert et al. (2020), Deng and Hooi (2021) are not considered. Last but not least, while they attempt to readapt the AUC using the recently introduced range-based metrics (Tatbul et al., 2018), they do not report the range-based precision, recall, and F1-score that are essential for an in-depth comparative study of existing methods, which is the core objective of the present paper. The work presented by Liu et al. (2022) compares DL techniques with and without the Point Adjustment protocol under different federated learning settings using classical metrics only. Similarly, Darban et al. (2022) provide a comprehensive review of DL-based anomaly detection for time-series, detailing fundamental principles, applications, and guidelines for practitioners. They compare several DL-based approaches using classical metrics but do not consider practical aspects like model size and stability. The study of Paparrizos et al. (2022) proposes a benchmark for evaluating univariate anomaly detection methods, mostly targeting classical ML approaches including only few DL methods and without adopting a unified protocol. Furthermore, Audibert et al. (2022) report a per-benchmark analysis between conventional, ML-based, and DL-based approaches, accounting for model stability but lacking range-based evaluation, per-anomaly type analysis, and a clear unified protocol. Lastly, Belay et al. (2023) focus on evaluating multivariate techniques without reporting any range-based performance, or per-anomaly type analysis.

**Table 1**
Comparison of existing evaluation studies of anomaly detection in time-series: we specify which of the following aspects were taken into account: ($i$) standard performance metrics which correspond to the precision, recall, and F1-score; ($ii$) revisited performance metrics extending the precision, recall, and F1-score to time-series introduced by Tatbul, Lee, Zdonik, Alam, and Gottschlich (2018); ($iii$) network size; ($iv$) consideration of ML approaches in the comparison; ($v$) evaluation of recent deep learning techniques; ($vi$) analysis with respect to the types of anomalies; and ($vii$) use of a unified experimental protocol. Note that by "partially" we mean that the authors briefly discussed the concept without necessarily producing any related comparison or results in their study.

| Papers | Analysis using standard performance metrics | Analysis using range-based metrics (Tatbul et al., 2018) | Network size | Eval. of recent DL techniques | Comparison against ML techniques | Analysis w.r.t anomaly types | Unified experimental protocol | Model stability |
|---|---|---|---|---|---|---|---|---|
| Choi, Yi, Park, and Yoon (2021) | yes | no | no | yes | no | no | no | no |
| Lai et al. (2021) | yes | no | no | no | yes | yes | yes | no |
| Wu and Keogh (2021) | no | no | no | no | no | no | no | no |
| Kim, Choi, Choi, Lee, and Yoon (2022) | yes | no | no | yes | no | no | yes | no |
| Schmidl, Wenig, and Papenbrock (2022) | yes | partially | partially | no | yes | yes | partially | no |
| Audibert, Michiardi, Guyard, Marti, and Zuluaga (2022) | yes | no | no | yes | yes | no | no | yes |
| Darban, Webb, Pan, Aggarwal, and Salehi (2022) | yes | no | no | yes | no | no | partially | no |
| Liu et al. (2022) | yes | no | partially | yes | no | no | partially | no |
| Paparrizos et al. (2022) | yes | partially | no | no | yes | yes | no | yes |
| Liu, Zhou, Yang, and Wang (2023) | no | no | partially | yes | no | no | no | no |
| Zhong et al. (2023) | no | partially | partially | yes | yes | no | partially | no |
| Belay, Blakseth, Rasheed, and Salvo Rossi (2023) | yes | no | partially | yes | yes | no | no | no |
| This work | yes | yes | yes | yes | yes | yes | yes | yes |

Hence, in this survey, we provide a comprehensive evaluation study of recent state-of-the-art algorithms by taking into account all the mentioned aspects (**i**) to (**v**). As summarized in Table 1, an analysis using standard performance metrics, as well as the novel performance metrics proposed by Tatbul et al. (2018) is performed. In addition, the number of parameters of DL-based approaches is reported as it directly impacts the memory consumption and the model scalability. Moreover, experiments according to the nature of anomalies are carried out using the taxonomy that was recently introduced by Lai et al. (2021). Lastly, a unified experimental protocol is used to compare existing methods. In short, this work aims to provide a comprehensive evaluation of numerous paradigm-representative time-series anomaly detection techniques, including recent deep learning methods, for a better assessment of their practical relevance. For that purpose, additional aspects are considered in complement to the traditional performance metrics, such as employing a unified experimental protocol, using range-based performance metrics, analyzing the performance based on the type of anomalies, and studying the model size and stability. The aim of this study is to help the community understand the advantages and limitations of state-of-the-art techniques from a broader applicative perspective and lay the foundations for better experimental evaluation practices.

The remainder of this paper is organized as follows. Section 2 presents preliminaries necessary for the understating of this paper. Section 3 reviews state-of-the-art time-series anomaly detection methods. Section 4 describes the used datasets and details the evaluation protocol considered in the experiments. Section 5 presents and analyzes the results. Finally, Section 6 concludes this work.

## 2. Preliminaries

A time-series is a temporally ordered set of $n$ variables which can be denoted by $X = \{X_t\}_{1 \leq t \leq N}$ where $X_t \in \mathbb{R}^n$ refers to the $n$-dimensional vector of variables at an instant $t$. Note that the time-series is univariate if $n = 1$, and is multivariate otherwise ($n > 1$). This section reviews the necessary background for a better understanding of this survey. Specifically, we start by recalling the different types of time-series anomalies according to the taxonomy of Lai et al. (2021). Then, we present the usual paradigms employed for anomaly detection in time-series.

### 2.1. Types of anomalies

As discussed by Choi et al. (2021), anomalies in time-series can generally be classified into three main categories, namely, *point*, *contextual*, and *collective* anomalies. However, unlike point anomalies, the definitions of contextual and collective ones are more ambiguous in the state-of-the-art, as stated by Lai et al. (2021). Indeed, they are heavily impacted by the application context. For instance, Yu, Zhu, Li, and Wan (2014) define contextual anomalies as small temporal segments formed by neighboring points, while Golmohammadi and Zaiane (2015) consider them as seasonal points (occurring periodically). Lai et al. have recently refined the definition of outlier types (Lai et al., 2021). They distinguish between *point-wise outliers* and *pattern-wise outliers*. The former is formed by *global* and *contextual* outliers while the latter is composed of *shapelet*, *seasonal*, and *trend* outliers. In the following, the taxonomy proposed by Lai et al. (2021), which is central to our analysis, is recalled.

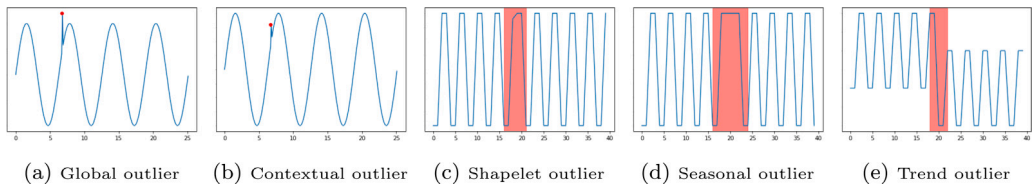

Fig. 1. Examples of the 5 different types of anomalies proposed by Lai et al. (2021).

*2.1.1. Point-wise anomalies*

Point-wise outliers are local anomalies occurring on individual time stamps. Let $X = \{X_t\}_{1 \leq t \leq N}$ be a multivariate time-series and $\hat{X}_t$ the expected value of $X_t$ at an instant $t$ according to a regression model. Given a well-chosen threshold $\delta > 0$, an anomaly at an instant $t$ can be formally defined by,

$$\|X_t - \hat{X}_t\| > \delta, \tag{1}$$

where $\|.\|$ defines an $L_p$ norm.

**Global outliers.** They can be seen as point-wise anomalies which importantly deviate from the rest of the points in a time-series. They usually correspond to spikes in the time-series, as shown in Fig. 1(a). In this case, the threshold $\delta$ can be formulated as,

$$\delta = \lambda \sigma(X), \tag{2}$$

where $\sigma(.)$ refers to the standard deviation operator and $\lambda \in \mathbb{R}^{+*}$.

**Contextual outliers.** They refer to individual points which differ significantly from their neighbors. The latter are often small glitches in the time-series as illustrated in Fig. 1(b). The threshold can be defined as,

$$\delta = \lambda \sigma(X_{t-k:t+k}), \tag{3}$$

where $X_{t-k:t+k} = \{X_{t-k}, X_{t-k+1}, \ldots, X_{t+k}\}$ is the signal corresponding to the temporal window centered on $t$. The function $\sigma(.)$ refers to the standard deviation operator and $\lambda \in \mathbb{R}^{+*}$.

*2.1.2. Pattern-wise anomalies*

Pattern-wise anomalies refer to anomalous sub-sequences which typically showcase discords or irregularities. These anomalies are defined by Lai et al. (2021) by modeling a time-series $X$ with spectral structural analysis (Granger & Watson, 1984) as follows,

$$X = \rho(2\pi\omega T) + \tau(T), \tag{4}$$

such that $\rho(2\pi\omega T) = \sum_k [A \sin(2\pi\omega_k T) + B \cos(2\pi\omega_k T)]$ corresponds to the base *shapelet* function which can be interpreted as the characteristic shape of $X$. The *seasonality*, which describes a pattern occurring at specific regular intervals in a time-series, is modeled with $\omega = \{w_1, w_2, \ldots, w_k\}$. Finally, a trend function denoted by $\tau$ defines the global direction of $X$. In particular, a sub-sequence $X_{i:j}$ of a time-series $X$ with $1 \leq i < j \leq N$ can be formulated using a shapelet function such that,

$$X_{i:j} = \rho(2\pi\omega T_{i,j}) + \tau(T_{i,j}), \tag{5}$$

with $\rho$, $\omega$, $\tau$, and $T_{i,j}$ respectively being the shape, the seasonality, the trend, and the time-stamps of the sub-sequence. The analysis of the shapelet, the seasonality as well as the trend functions allow distinguishing the three following outliers:

**Shapelet outliers.** They represent the anomalous sub-sequences enclosing shapelets that are different from the expected ones, as shown in Fig. 1(c). The following condition can be used to define shapelet outliers as follows,

$$d_\rho(\rho(.), \hat{\rho}(.)) > \delta, \tag{6}$$

with $d_\rho$ being a dissimilarity measure computed between two sets of shapelets. $\hat{\rho}(.)$ corresponds to the expected shapelets in a given sub-sequence and $\delta$ is the threshold.

**Seasonal outliers.** They can be defined as sub-sequences with unexpected seasonalities with respect to the full sequence, as illustrated in Fig. 1(d).

$$d_\omega(\omega, \hat{\omega}) > \delta, \tag{7}$$

with $d_\omega$ being a dissimilarity measure between two seasonality, $\hat{\omega}$ being the expected seasonality in the sub-sequence, and $\delta$ being the threshold.

**Trend outliers.** They refer to sub-sequences with an importantly altered trend. Consequently, a shift in the mean data can be observed, as shown in Fig. 1(e). Mathematically, trend outliers can be defined by,

$$d_\tau(\tau, \hat{\tau}) > \delta \tag{8}$$

where $d_\tau$ is a dissimilarity measure computed between two trends, $\hat{\tau}$ is the expected trend of the sub-sequence, and $\delta$ is the threshold.

*2.2. Paradigms for anomaly detection in times-series*

Existing anomaly detection methods in time-series mainly employ five different paradigms, namely, clustering-based, density estimation-based, distance-based, reconstruction-based and forecasting-based methods (see Fig. 2).

*2.2.1. Clustering-based methods*

Let $S^n$ be the feature space of multivariate time-series of dimension $n$. Let $\mathcal{N}^n$ be the estimated sub-space of normal time-series of dimension $n$ such that $\mathcal{N}^n \subset S^n$. Let $f$ be a feature extractor function which maps an input time-series $X \in \mathbb{R}^{n \times N}$ to $S^n$. An anomaly is detected if,

$$f(X) \notin \mathcal{N}^n. \tag{9}$$

Note that the classification of $X$ as an anomaly or not can also be determined with the use of a distance that is compared to a threshold. This is the case, for example, of Support Vector Data Description (SVDD) (Tax & Duin, 2004), which measures the distance from the centroids.

*2.2.2. Density estimation-based methods*

Density estimation-based methods mainly aim to estimate the probability density function of normal time-series denoted as $\mathbf{p}_\theta$. Given a time-series $X$, the likelihood function $\mathcal{L}$ of $\theta$ and a threshold $\tau$, an anomaly is detected if,

$$\mathcal{L}(\theta|X) > \tau. \tag{10}$$

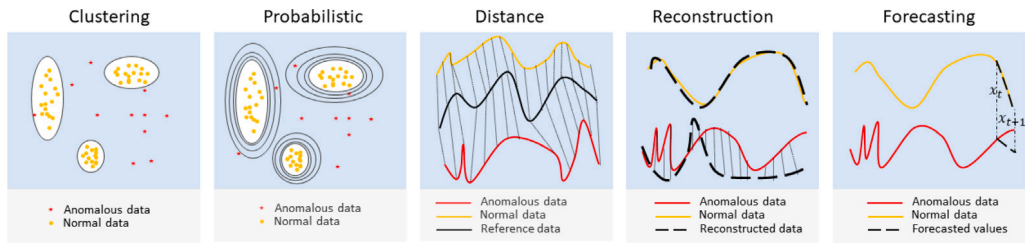

**Fig. 2.** Overview of the different paradigms for anomaly detection in time-series: in contrast to clustering and probabilistic approaches, distance-based, reconstruction-based and forecasting-based approaches take into account the temporal aspect.

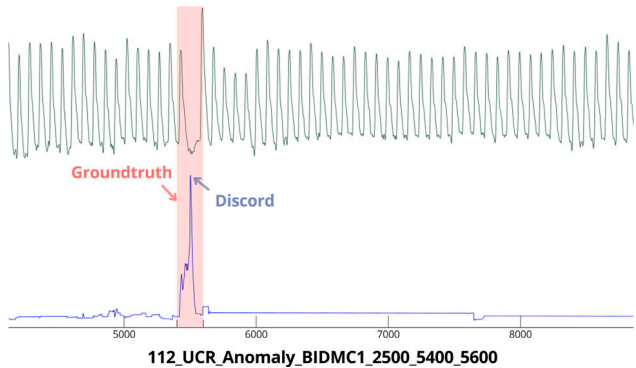

**Fig. 3.** An example of time-series from the UCR dataset, where the discord was calculated using DAMP (Lu, Wu, Mueen, Zuluaga, & Keogh, 2022).

#### 2.2.3. Distance-based methods

Distance-based methods rely on the definition of an adequate distance between two temporal sequences. This distance should measure the dissimilarity between them. Let $X$ and $R$ be, respectively, a given time-series and a reference normal time-series. Let us denote by $D$ a distance for time-series. Given a predefined threshold $\delta$, an anomaly is detected in $X$ if,

$$D(X, R) > \delta. \quad (11)$$

#### 2.2.4. Reconstruction-based methods

Reconstruction-based approaches aim at learning a model for the accurate and full reconstruction of a normal time-series. The assumption is that the learned model will fail when reconstructing abnormal sequences. Let $X$ and $\hat{X}$ be respectively the original and the reconstructed time-series. Given a predefined threshold $\delta$, an anomaly is detected in $X$ if,

$$\|X - \hat{X}\| > \delta. \quad (12)$$

#### 2.2.5. Forecasting-based methods

Forecasting-based approaches are based on the prediction of future states given previous observations. Similar to reconstruction-based methods, they assume that the prediction will be less accurate in the presence of an anomaly. Let $X = \{X_0, X_1, \ldots, X_N\}$ be a time-series where $X_i$ refers to an observation of $X$ at an instant $i$. Given a threshold $\delta$, an anomaly is detected at an instant $i$ if,

$$\|\hat{X}_i - X_i\| > \delta, \quad (13)$$

where $\hat{X}_{t_i}$ corresponds to the predicted state given the observation $X_{0:i-1} = \{X_0, X_1, \ldots, X_{i-1}\}$.

### 3. State-of-the-art on time-series anomaly detection

Over the last two decades, the research community has widely explored the field of anomaly detection (Han, Hu, Huang, Jiang, & Zhao, 2022), including anomaly detection in time-series. The latter can be addressed from five different perspectives. As reported in Section 2, we distinguish between clustering-based, density-estimation-based, distance-based, reconstruction-based, and forecasting-based techniques. Earlier techniques have investigated these five different paradigms by exploiting traditional machine learning (Jin et al., 2019; Liu et al., 2008) and statistical tools (Yaacob et al., 2010). Nevertheless, as mentioned in Choi et al. (2021), these approaches have shown a drop in performance when dealing with high-dimensional time series. Given the recent advances in DL, DNNs have been considered as an alternative (Audibert et al., 2020; Li et al., 2019; Malhotra et al., 2015; Su et al., 2019), mainly taking inspiration from traditional methods. In the following, we review these five categories of approaches, starting with conventional techniques (see Fig. 3), then moving to current DL methods.

#### 3.1. Clustering-based methods

Clustering-based methods are discriminative approaches aiming to estimate explicitly or implicitly decision boundaries for detecting anomalies (Liu et al., 2008; Schölkopf, Williamson, Smola, Shawe-Taylor, & Platt, 1999; Tax & Duin, 2004) as depicted in Eq. (9). One-Class Support Vector Machine (OC-SVM) (Schölkopf et al., 1999) is probably one of the most popular algorithms for anomaly detection. Its goal is to estimate the support of a high-dimensional distribution. This one-class classification method has been mainly used for detecting time-independent anomalies (Li, Huang, Tian, & Xu, 2003; Wang, Wong, & Miner, 2004) but has also been employed for isolating outliers in time-series (Ma & Perkins, 2003). Inspired by Support Vector Machines (SVM), Support Vector Data Description (SVDD) (Tax & Duin, 2004) is another well-known method that is often used in the context of anomaly detection (Huang et al., 2017). Similar to SVM, kernels that map data representations to a higher dimensional space can be used. However, instead of relying on the estimation of a hyperplane, SVDD computes spherically shaped boundaries.

Shallow clustering-based approaches necessitate the hand-crafting of discriminative features and often require the selection of an appropriate kernel. Recently, with the advances in DL, there have been attempts to extend these classical approaches. Most of these methods, such as Deep SVDD (Ruff et al., 2018) variants, extend traditional methods by learning a kernel that maps data to a discriminative high-dimensional feature space. This is usually carried out by optimizing a Neural Network. These approaches have shown promising results when dealing with non-sequential data. Unfortunately, the temporal modeling of time-series is often disregarded, mainly relying on a simple sliding window. As a solution, Shen, Li, and Kwok (2020) suggest fusing multi-scale temporal features and employing a Recurrent Neural Networks (RNN) to model temporal dependencies.

#### 3.2. Density-estimation methods

As described in Section 2.2.2, these probabilistic approaches detect anomalies by estimating the normal data density function. For example, Breunig, Kriegel, Ng, and Sander (2000) proposed a method referred to as Local Outlier Factor (LOF) to detect anomalies by computing the local density. Tang, Chen, Fu, and Cheung (2002) calculate the local connectivity for determining anomalies instead. In Yairi et al. (2017) and Lindstrom, Jung, and Larocque (2020), a Gaussian Mixture Model (GMM) and Kernel Density Estimation (KDE) are respectively used for estimating the density of normal representations. Over the last years, efforts have been made to introduce DNN-based probabilistic methods. For instance, Zong et al. (2018) proposed to train an auto-encoder for extracting relevant representations before fitting a GMM. Nevertheless, as for clustering-based methods, probabilistic approaches usually do not model the temporal aspect restricting their effectiveness in the context of time-series anomaly detection.

### 3.3. Distance-based methods

Distance-based methods usually define explicitly a distance between a time-series and a reference to detect anomalies (Baptista, Demisse, Aouada, & Ottersten, 2018; Baptista et al., 2019; Benkabou, Benabdeslem, & Canitia, 2018; Diab et al., 2019), as described in Section 2.2.3. Among the most used distance-based algorithm, one can refer to Dynamic Time Warping (Berndt & Clifford, 1994), which aims at finding the optimal match between two ordered sequences. Earlier distance-based methods are mostly characterized by a relatively high complexity induced by the optimal matching and the need for defining a reference time-series (Baptista et al., 2018). To address those issues, some methods such as Hundman, Constantinou, Laporte, Colwell, and Soderstrom (2018) reduce the computational cost by only using a small initial snippet instead of a full reference. Alternatively, the DAMP algorithm introduced by Lu et al. (2022) can efficiently handle datasets with trillions of data points, by implementing strategies like iterative doubling for backward nearest neighbor search, forward processing for pruning non-discord subsequences, and relying on parallel vectors to reduce the computation cost.

### 3.4. Reconstruction-based methods

Reconstruction-based methods aim at reconstructing the entire time-series, as presented in Section 2.2.4. Shallow reconstruction-based time-series anomaly detection methods (Hyndman, Wang, & Laptev, 2015; Jin, Qiu, Sun, Peng, & Zhou, 2017; Wang, Miranda-Moreno, & Sun, 2021) have mainly adopted Principal Component Analysis (Kirby & Sirovich, 1990) (PCA) or its variants such as kernel PCA (kPCA) (Schölkopf, Smola, & Müller, 1998). These approaches estimate an orthogonal projection, then compute a reconstruction error between the original and reconstructed time-series. Lately, Auto-Encoders (AE) (Oja, 1982) have been introduced as the deep learning-based counterpart of PCA. Unsurprisingly, the latter has been adopted in the context of anomaly detection in time-series (Reddy, Sarkar, Venugopalan, & Giering, 2016). For example, Lin et al. (2020) introduce a Long-Short Term Memory Variation Auto-Encoder (LSTM-VAE) architecture. While the Variation Auto-Encoder architecture (VAE) is used for learning robust representations, a Long Short-Term Memory (LSTM) network allows modeling temporal dependencies. Generative Adversarial Networks (GAN) have also been proposed as a reconstruction-based method. In Audibert et al. (2020), Audibert et al. attempted to take the best of both worlds. In particular, they introduced adversarially trained autoencoders for detecting anomalies in time-series.

### 3.5. Forecasting-based methods

As discussed in Section 2.2.5, traditional forecasting-based anomaly detection methods are primarily based on auto-regression-based models such as AutoRegressive Integrated Moving Average (ARIMA) (Moayedi & Masnadi-Shirazi, 2008). With the recent advances in deep learning, LSTM has been used to replace auto-regression models (Malhotra et al., 2015). This architecture allows modeling short-term as well as long-term temporal dependencies. Deng and Hooi (2021) have recently proposed a graph-based deep learning model with an attention mechanism for capturing multivariate correlations.

### 3.6. Hybrid methods

As discussed by Zhao et al. (2020), reconstruction and forecasting-based approaches have shown to be, so far, the best candidates for anomaly detection in time-series. While reconstruction-based methods allow modeling inconsistencies within the global distribution of time series, forecasting-based approaches are more appropriate for capturing local anomalies. For that reason, Zhao et al. (2020) have introduced a hybrid method leveraging these two complementary paradigms. Specifically, they design a two-stream attention-based graph network that simultaneously optimizes forecasting and reconstruction losses.

## 4. Datasets and evaluation protocol

In this section, the datasets, the evaluation criteria, the pre-processing and post-processing algorithms as well as the considered methods for the experiments are presented.

### 4.1. Datasets

A total of five datasets have been considered for evaluating recent methods for anomaly detection in time-series. Table 2 details the different characteristics of each dataset. The considered benchmarks are:

**Secure Water Treatment (SWaT).** It is a dataset collected from a testbed water treatment for 11 days. During the last 4 days, 36 attacks of different duration and natures have been introduced. The data collected over the 7 first days have been used for training in all our experiments. During this period, the water treatment was carried out under normal conditions. In contrast, the data gathered during the last 4 days were exposed to multiple attacks. The latter have been used for testing[1].

**Mars Science Laboratory (MSL).** It is formed by 27 telemetry signals collected from the Curiosity Rover spacecraft on Mars. Each signal consists of a multivariate time-series of dimension 55. The first dimension encloses telemetry data, while the remaining 54 correspond to a one-hot encoded command. The publicly available dataset has been released by NASA (O'Neill, Entekhabi, Njoku, & Kellogg, 2010). The training and testing data are separated, and anomalies are annotated. Nevertheless, it can be noted that the experimental protocol varies from one reference to another. In particular, some studies such as Tinawi (2019) ignore the one-hot encoded vector considering only telemetry data. In addition, other approaches such as Deng and Hooi (2021) combine the telemetric data from 27 signals assuming that it forms a unique dataset. Nevertheless, in most cases, authors do not provide sufficient information about their experimental protocol, making a direct comparison not straightforward. In this paper, we follow the experimental protocol of Hundman et al. (2018). Each signal is considered to be a separate and independent multivariate sub-dataset. This means that the training and testing phases are performed each time on one single sub-dataset. Finally, the average performance is reported.

**Soil Moisture Active Passive dataset (SMAP).** This dataset contains telemetry data and one-hot encoded vectors similar to the MSL dataset. It has also been released by NASA (O'Neill et al., 2010). However, in this case, the dataset is formed by 53 signals received from the Soil Moisture Active Passive satellite. The annotated training and testing data are provided. Nevertheless, as for the MSL dataset, similar inconsistencies regarding the experimental protocol can be remarked. For that reason, we propose using the protocol of Hundman et al. (2018), where each signal is considered to be a separate and independent multivariate sub-dataset. This leads to train and test on 53 different sub-datasets and reporting the obtained average performance.

---

[1] Details of the SWaT dataset can be found here.

Table 2
Summary of the 5 datasets used in the experiments. The percentage of anomalies in the testing set is reported.

|  | SWaT | MSL | SMAP | UCR | TODS |
| --- | --- | --- | --- | --- | --- |
| Number of datasets | 1 | 27 | 55 | 250 | 5 |
| Variables | 52 | 55 | 25 | 1 | 10 |
| Percentage of anomalies | 12.14 | 10.48 | 12.82 | 0.38 | 5 |
| Training data points | 495000 | 58317 | 138004 | 5302449 | 10000 |
| Testing data points | 449919 | 73729 | 435826 | 12919799 | 10000 |
| Type of data | Real | Real | Real | Real | Synthetic |
| Type of anomalies | Artificially forced | Natural | Natural | Natural/Synthetic | Synthetic |

Table 3
Paradigm type and nature of evaluated methods.

| Method | Type of paradigm | Nature |
| --- | --- | --- |
| OC-SVM (Schölkopf et al., 1999) | Clustering | Shallow |
| iForest (Liu et al., 2008) | Clustering | Shallow |
| ARIMA (Moayedi & Masnadi-Shirazi, 2008) | Forecasting | Shallow |
| DAMP (Lu et al., 2022) | Distance-based | Shallow |
| DA-GMM (Zong et al., 2018) | Density-estimation | Deep |
| THOC (Shen et al., 2020) | Clustering | Deep |
| USAD (Audibert et al., 2020) | Reconstruction | Deep |
| GDN (Deng & Hooi, 2021) | Forecasting | Deep |
| MTAD-GAT (Zhao et al., 2020) | Hybrid (Forecasting & Reconstruction) | Deep |

**UCR time series anomaly archive (UCR).** It has been recently proposed by Wu and Keogh (2021). In this work, the authors claim that most of the existing anomaly detection datasets are *trivial*. By trivial, they mean that an anomaly can be detected with a single line of MATLAB code. They also criticize the lack of realism and annotation precision in current datasets. As an alternative, they introduce the UCR dataset, which gathers 250 realistic sub-datasets. This dataset is collected from various fields, including medicine, sports, and robotics. The training and test sets are well-defined.[2]

**Automated Time-series Outlier Detection System (TODS).** It is a collection of 5 synthetically generated multivariate datasets. The dataset was generated using the source code from Lai et al. (2021), therefore producing different types of anomalies following the taxonomy of Lai et al. (2021). The dimension of the generated time-series is 10. Training datasets contain only normal values, while testing datasets incorporate 5 different types of anomalies. The annotation of the outlier types is provided, therefore allowing a per-type analysis.

*4.2. Evaluation criteria*

In this section, we present the used evaluation criteria. **Precision, recall and F1-scores.** The most common metrics used to evaluate the performance of time-series anomaly detection algorithms are the precision computed as follows,

$$\text{Precision} = \frac{\text{True positives}}{\text{True positives} + \text{False positives}}, \quad (14)$$

the recall is calculated as below,

$$\text{Recall} = \frac{\text{True positives}}{\text{True Positives} + \text{False Negatives}} \quad (15)$$

and the F1-score corresponding to,

$$\text{F1-score} = \frac{2 \cdot \text{Precision} \cdot \text{Recall}}{\text{Precision} + \text{Recall}}. \quad (16)$$

It is worth noting that some methods such as the work of Wu and Keogh, 2022, generally opt for other evaluation metrics. For instance,

---
[2] More information about the UCR dataset can be found here.

in Wu and Keogh (2021) the authors argue that time-series datasets should only have a single anomalous sequence per series. Under such a setting they propose using accuracy to assess whether an anomaly has been correctly identified. Such a binary score is often simple to interpret and to use and can be used for introducing more flexibility. On the other hand, standard methods provide a more comprehensive and explainable assessment of performance across multiple instances, namely true positives, false positives, and false negatives. They can handle more than a single anomaly subsequence in a given time-series, and are therefore adapted to data with different anomaly ratios such as most of the considered benchmarks (SWaT, MSL, SMAP).

**Revisited precision, recall and F1-scores for time-series.** In addition to conventional performance metrics, more recent and elaborate performance metrics tailored to time-series introduced by Tatbul et al. (2018) are considered. These metrics extend classical precision, recall, and F1-score, from point-based to range-based anomaly detection. Fig. 4 highlights the distinction between point-based and range-based anomalies. Contrary to the case of point-based approaches, a prediction in a time-series can be both a true positive (TP) and a false negative (FN) due to *partial overlap* with the ground-truth as shown in Fig. 4b. Therefore, as discussed by Tatbul et al. (2018), a more informative time-series evaluation process should (1) quantify the size of the partial overlap; (2) identify the overlap position, and; (3) take into account its cardinality, i.e., with how many anomalous ground-truth sub-sequences it overlaps. More specifically, given a set of real anomaly sequences $R = \{R_1, \ldots, R_{N_r}\}$ and a set of predicted anomaly sequences $P = \{P_1, \ldots, P_{N_p}\}$ the recall is expressed with respect to the number of real anomalies $N_r$ in a dataset (Tatbul et al., 2018). It seeks to reward a detector when it predicts a TP and penalizes it when the prediction is an FN as follows,

$$\text{Recall}_T(R, P) = \frac{1}{N_r} \sum_{i=1}^{N_r} \text{Recall}_T(R_i, P), \quad (17)$$

and,

$$\text{Recall}_T(R_i, P) = \alpha \cdot \mathbb{1}_{\sum_{j=1}^{N_p} |R_i \cap P_j| \geq 1} + \frac{1-\alpha}{\sum_{j=1}^{N_p} |R_i \cap P_j|} \cdot S_c(R_i, P), \quad (18)$$

where $0 \leq \alpha \leq 1$ is a scaling factor that rewards the detector when it detects the existence of the anomaly $R_i$ and $\mathbb{1}$ is an indicator function. Finally, $S_c(R_i, P)$ which quantifies the overlap size is computed based on the cumulative overlap size $\omega$ as follows,

$$S_c(R_i, P) = \sum_{j=1}^{N_p} \omega(R_i, R_i \cap P_j, \delta), \quad (19)$$

where $\delta$ returns a score depending on the overlap location between $R_i$ and a prediction $P_j$ (flat bias, front bias, middle bias, and back bias). Further details could be found in the original manuscript of Tatbul et al. (2018). The precision is similarly defined. It seeks to assess the quality of the predictions by rewarding a detector in the presence of a TP and penalizing it when facing an FP. It is computed as follows,

$$\text{Precision}_T = \frac{1}{N_p} \sum_{i=1}^{N_p} \text{Precision}_T(R, P_i), \quad (20)$$

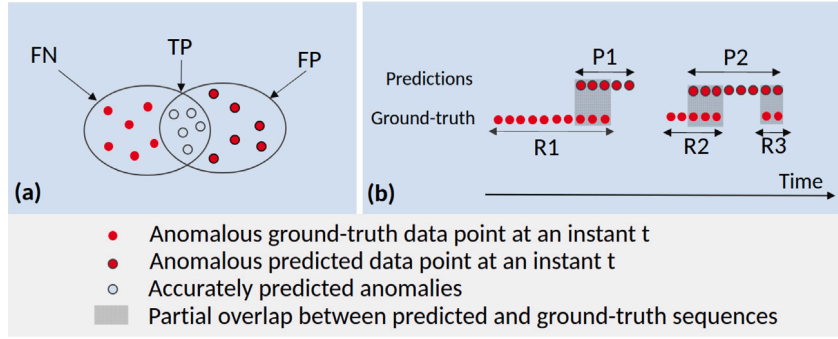

**Fig. 4.** The evaluation process of (a) point-based anomalies versus (b) range-based anomalies. Range-based anomalies are characterized by partial overlap(s) with the ground-truth. A more accurate evaluation for time-series should quantify the overlap in terms of **size**, **position**, and **cardinality**.

and,

$$\text{Precision}_T(R, P_i) = \frac{1}{\sum_{j=1}^{N_r} |R_j \cap P_i|} \cdot S(R, P_i), \quad (21)$$

where $S_c(R, P_i)$ quantifies the cumulative overlap between the considered prediction $P_i$ and all the ground-truths in $R$ as explained in Eq. (19). It is expressed as,

$$S_c(R, P_i) = \sum_{j=1}^{N_r} \omega(P_i, P_i \cap R_j, \delta). \quad (22)$$

Finally, the F1-score is redefined as follows,

$$\text{F1-score}_T = \frac{2 \cdot \text{Precision}_T \cdot \text{Recall}_T}{\text{Precision}_T + \text{Recall}_T} \quad (23)$$

**Model stability.** We define model stability as the ability of a machine/deep learning algorithm to reproduce similar results when retrained under the same conditions. While OC-SVM ensures stability because of its deterministic nature, most of the considered methods rely on a random parameter initialization which may impact the final performance of the model. Ideally, the model should achieve the same results regardless of this initialization. To assess the stability, each experiment is carried out five times. Then, the mean and standard deviation of those five runs are reported. A lower standard deviation reflects higher stability. To the best of our knowledge, we are among the first to analyze this aspect experimentally in the context of anomaly detection in time-series.

**Generalization to different types of anomalies.** We propose reporting the performance according to the anomaly type encountered. This analysis can help identify the most suitable algorithm for a given application. To that aim, the TODS benchmark, which encloses the annotation of 5 different types of outliers depicted in Section 4.1 is used. Although the definition of different anomaly types has been reported in several references, very few works have carried out an experimental study with respect to the anomaly type. A rare example we can mention is the work of Lai et al. (2021). Nevertheless, it can be noted that in this paper, recent DL-based state-of-the-art approaches such as GDN (Deng & Hooi, 2021), USAD (Audibert et al., 2020), and MTAD-GAT (Zhao et al., 2020) are not evaluated. For each anomaly type, the percentage of well-detected anomalies is reported.

**Model size.** In a real-world context, deploying algorithms on specific hardware with a limited memory capacity can be challenging. Therefore, being aware of the model size, which directly impacts the memory consumption, is a crucial component often neglected. For that purpose, we report the number of parameters and the size in MegaBytes (MB) of the trained deep learning models considered for this evaluation.

*4.3. Post-processing and pre-processing*

**Data Normalization.** Normalizing the data is a common practice in machine learning, particularly in anomaly detection. Hence, for the sake of fairness, a data normalization pre-processing was applied in all our experiments. More specifically, the data are normalized using the maximum and minimum values in the training data as in Zhao et al. (2020).

**Point Adjustment.** Point adjustment initially introduced by Xu et al. (2018) is a protocol that adjusts the predictions before computing performance metrics. It acts as follows: if at least one point is classified as an anomaly in an outlier segment, all the predictions in that segment are set to anomalous. The idea behind this protocol is that an algorithm triggering an alert for any point in a contiguous anomaly segment might be sufficient for a timely reaction. Fig. 5 illustrates the point adjustment protocol by showing the ground-truth, the original predictions, and the predictions after point adjustment of a given time-series. After applying the point adjustment protocol, the F1-Score goes from 0.32 to 0.85. This significant gap has therefore raised some concerns in the literature regarding the use of point adjustment. For example, Kim et al. (2022) claim that by using this protocol, a randomly generated anomaly score might outperform several recently proposed time-series anomaly detection algorithms. In this paper, we report the performance of existing methods with and without point adjustment.

**Fig. 5.** Application of the Point Adjustment (PA) on a given time-series: the Ground-Truth (GT), the original prediction (Pred) and the prediction after Point Adjustment (PA) are reported. In this example, the performance of the algorithm without and with point adjustment is respectively: Precision = 0.75, Recall = 0.2, F1-Score = 0.32, and Precision = 0.92, Recall = 0.79, F1-Score = 0.85. Best viewed in colors.

*4.4. Evaluated methods*

We consider in total nine anomaly detection methods. Table 3 summarizes the characteristics of each evaluated method.

Four shallow standard methods are evaluated, namely, OC-SVM (Schölkopf et al., 1999), iForest (Liu et al., 2008), ARIMA (Moayedi & Masnadi-Shirazi, 2008) and DAMP (Lu et al., 2022). In addition, five recent DL-based methods have been considered: DA-GMM (Zong et al., 2018), THOC (Shen et al., 2020) USAD (Audibert et al., 2020), GDN (Deng & Hooi, 2021) and MTAD-GAT (Zhao et al., 2020). The latter has been selected according to the following criteria: (1) Relevance of the topic: all the chosen anomaly detection algorithms are unsupervised and have been specifically designed for detecting anomalies in time-series. (2) Publication date: all the DL-based algorithms

**Table 4**

Results in terms of traditional performance metrics of evaluated state-of-the-art methods (precision P, recall R, F1-score) on the 5 considered datasets **without Point Adjustment (PA)**. The experiments have been performed 5 times for each algorithm and dataset. The mean and standard deviation are reported. The bold and underlined results correspond to the first and second-best F1-Score respectively.

|          |    | SWaT          | MSL              | SMAP             | UCR              | TODS             | Avg. method F1   |
|----------|----|---------------|------------------|------------------|------------------|------------------|------------------|
| USAD     | P  | 0.28 ± 0.02   | 0.15 ± 0.01      | 0.18 ± 0.01      | 0.01 ± 0.00      | 0.05 ± 0.00      |                  |
|          | R  | 0.74 ± 0.01   | 0.57 ± 0.05      | 0.49 ± 0.01      | 0.48 ± 0.00      | 0.54 ± 0.02      |                  |
|          | F1 | 0.41 ± 0.02   | 0.21 ± 0.02      | 0.21 ± 0.01      | 0.02 ± 0.00      | 0.10 ± 0.00      | 0.19 ± 0.13      |
| GDN      | P  | 0.34 ± 0.03   | 0.31 ± 0.01      | 0.25 ± 0.00      | 0.12 ± 0.00      | 0.07 ± 0.01      |                  |
|          | R  | 0.72 ± 0.04   | 0.64 ± 0.02      | 0.55 ± 0.04      | 0.42 ± 0.00      | 0.59 ± 0.16      |                  |
|          | F1 | 0.46 ± 0.03   | <u>0.35 ± 0.01</u> | <u>0.33 ± 0.01</u> | 0.12 ± 0.00      | <u>0.11 ± 0.00</u> | 0.27 ± 0.14      |
| THOC     | P  | 0.62 ± 0.16   | 0.22 ± 0.01      | 0.16 ± 0.01      | 0.01 ± 0.01      | 0.05 ± 0.01      |                  |
|          | R  | 0.46 ± 0.13   | 0.46 ± 0.02      | 0.27 ± 0.01      | 0.00 ± 0.01      | 0.19 ± 0.03      |                  |
|          | F1 | 0.52 ± 0.14   | 0.25 ± 0.01      | 0.12 ± 0.01      | 0.00 ± 0.00      | 0.08 ± 0.14      | 0.19 ± 0.18      |
| MTAD-GAT | P  | 0.85 ± 0.04   | 0.57 ± 0.04      | 0.58 ± 0.03      | 0.10 ± 0.00      | 0.16 ± 0.08      |                  |
|          | R  | 0.90 ± 0.03   | 0.79 ± 0.03      | 0.87 ± 0.03      | 0.28 ± 0.01      | 0.01 ± 0.02      |                  |
|          | F1 | **0.87 ± 0.01** | **0.60 ± 0.03** | **0.65 ± 0.03** | <u>0.13 ± 0.01</u> | 0.02 ± 0.03      | **0.45 ± 0.32**  |
| DAGMM    | P  | 0.43 ± 0.00   | 0.12 ± 0.01      | 0.11 ± 0.01      | 0.01 ± 0.00      | 0.12 ± 0.00      |                  |
|          | R  | 0.71 ± 0.00   | 0.19 ± 0.00      | 0.17 ± 0.00      | 0.20 ± 0.00      | 0.49 ± 0.00      |                  |
|          | F1 | <u>0.54 ± 0.00</u> | 0.12 ± 0.00 | 0.10 ± 0.01      | 0.01 ± 0.00      | **0.19 ± 0.00**  | 0.19 ± 0.18      |
| OCSVM    | P  | 0.24 ± 0.00   | 0.15 ± 0.00      | 0.12 ± 0.00      | 0.01 ± 0.00      | 0.05 ± 0.00      |                  |
|          | R  | 0.85 ± 0.00   | 0.66 ± 0.00      | 0.66 ± 0.00      | 0.73 ± 0.00      | 0.85 ± 0.00      |                  |
|          | F1 | 0.37 ± 0.00   | 0.24 ± 0.00      | 0.20 ± 0.00      | 0.02 ± 0.00      | 0.09 ± 0.00      | 0.18 ± 0.12      |
| iForest  | P  | 0.23 ± 0.10   | 0.18 ± 0.04      | 0.10 ± 0.01      | 0.05 ± 0.01      | 0.05 ± 0.01      |                  |
|          | R  | 0.83 ± 0.10   | 0.16 ± 0.05      | 0.04 ± 0.01      | 0.12 ± 0.01      | 0.04 ± 0.01      |                  |
|          | F1 | 0.36 ± 0.10   | 0.17 ± 0.03      | 0.08 ± 0.00      | 0.07 ± 0.00      | 0.04 ± 0.01      | 0.14 ± 0.12      |
| ARIMA    | P  | 0.13 ± 0.00   | 0.28 ± 0.00      | 0.17 ± 0.00      | 0.01 ± 0.00      | 0.05 ± 0.00      |                  |
|          | R  | 0.99 ± 0.00   | 0.83 ± 0.00      | 0.82 ± 0.00      | 0.85 ± 0.00      | 0.69 ± 0.00      |                  |
|          | F1 | 0.23 ± 0.00   | 0.28 ± 0.00      | 0.19 ± 0.00      | 0.02 ± 0.00      | 0.09 ± 0.00      | 0.16 ± 0.09      |
| DAMP     | P  | –             | –                | –                | 0.33 ± 0.00      | –                | –                |
|          | R  | –             | –                | –                | 0.34 ± 0.00      | –                | –                |
|          | F1 | –             | –                | –                | **0.28 ± 0.00**  | –                | –                |

**Table 5**

Results in terms of traditional performance metrics of evaluated state-of-the-art methods (precision (P), recall (R), F1-score (F1)) on the 5 considered datasets **with Point Adjustment (PA)**. The experiments have been performed 5 times for each algorithm and dataset. The mean and standard deviation are reported. The bold and underlined results correspond to the first and second-best F1-Score respectively.

|          |    | SWaT              | MSL              | SMAP             | UCR              | TODS             | Avg. method F1  |
|----------|----|-------------------|------------------|------------------|------------------|------------------|-----------------|
| USAD     | P  | 0.32 ± 0.02       | 0.22 ± 0.02      | 0.26 ± 0.01      | 0.02 ± 0.00      | 0.07 ± 0.00      |                 |
|          | R  | 0.89 ± 0.03       | 0.99 ± 0.03      | 0.95 ± 0.01      | 0.95 ± 0.00      | 0.71 ± 0.02      |                 |
|          | F1 | 0.47 ± 0.02       | 0.33 ± 0.01      | 0.34 ± 0.02      | 0.04 ± 0.00      | 0.13 ± 0.00      | 0.26 ± 0.16     |
| GDN      | P  | 0.40 ± 0.05       | 0.39 ± 0.02      | 0.36 ± 0.01      | 0.31 ± 0.01      | 0.10 ± 0.02      |                 |
|          | R  | 0.72 ± 0.04       | 1.00 ± 0.00      | 1.00 ± 0.00      | 0.99 ± 0.00      | 0.75 ± 0.12      |                 |
|          | F1 | 0.57 ± 0.05       | 0.50 ± 0.02      | <u>0.46 ± 0.01</u> | <u>0.39 ± 0.01</u> | 0.16 ± 0.02    | 0.42 ± 0.14     |
| THOC     | P  | 0.77 ± 0.08       | 0.31 ± 0.01      | 0.26 ± 0.01      | 0.06 ± 0.02      | 0.09 ± 0.01      |                 |
|          | R  | 0.86 ± 0.02       | 0.87 ± 0.02      | 0.84 ± 0.02      | 0.06 ± 0.02      | 0.35 ± 0.06      |                 |
|          | F1 | <u>0.81 ± 0.05</u> | 0.41 ± 0.02    | 0.34 ± 0.01      | 0.06 ± 0.02      | 0.14 ± 0.02      | 0.35 ± 0.26     |
| MTAD-GAT | P  | 0.86 ± 0.04       | 0.60 ± 0.04      | 0.59 ± 0.03      | 0.17 ± 0.00      | 0.16 ± 0.08      |                 |
|          | R  | 0.96 ± 0.03       | 0.86 ± 0.03      | 0.91 ± 0.03      | 0.57 ± 0.02      | 0.01 ± 0.02      |                 |
|          | F1 | **0.90 ± 0.01**   | **0.64 ± 0.04**  | **0.67 ± 0.03**  | 0.25 ± 0.01      | 0.02 ± 0.03      | **0.50 ± 0.32** |
| DAGMM    | P  | 0.49 ± 0.00       | 0.20 ± 0.00      | 0.16 ± 0.00      | 0.03 ± 0.00      | 0.15 ± 0.00      |                 |
|          | R  | 0.90 ± 0.00       | 0.44 ± 0.00      | 0.41 ± 0.00      | 0.78 ± 0.00      | 0.63 ± 0.00      |                 |
|          | F1 | 0.64 ± 0.00       | 0.25 ± 0.00      | 0.19 ± 0.00      | 0.06 ± 0.00      | **0.24 ± 0.00**  | 0.28 ± 0.19     |
| OCSVM    | P  | 0.26 ± 0.00       | 0.25 ± 0.00      | 0.15 ± 0.00      | 0.02 ± 0.00      | 0.05 ± 0.00      |                 |
|          | R  | 0.95 ± 0.00       | 0.95 ± 0.00      | 0.85 ± 0.00      | 0.85 ± 0.00      | 0.85 ± 0.00      |                 |
|          | F1 | 0.41 ± 0.00       | 0.40 ± 0.00      | 0.26 ± 0.00      | 0.04 ± 0.00      | 0.09 ± 0.00      | 0.24 ± 0.15     |
| iForest  | P  | 0.26 ± 0.12       | 0.47 ± 0.04      | 0.10 ± 0.01      | 0.16 ± 0.01      | 0.17 ± 0.06      |                 |
|          | R  | 0.97 ± 0.00       | 0.66 ± 0.06      | 0.04 ± 0.01      | 0.45 ± 0.01      | 0.17 ± 0.02      |                 |
|          | F1 | 0.40 ± 0.13       | <u>0.55 ± 0.04</u> | 0.36 ± 0.01    | 0.24 ± 0.01      | <u>0.17 ± 0.02</u> | 0.34 ± 0.13   |
| ARIMA    | P  | 0.13 ± 0.00       | 0.31 ± 0.00      | 0.18 ± 0.00      | 0.01 ± 0.00      | 0.06 ± 0.00      |                 |
|          | R  | 1.00 ± 0.00       | 1.00 ± 0.00      | 0.96 ± 0.00      | 0.97 ± 0.00      | 0.87 ± 0.00      |                 |
|          | F1 | 0.23 ± 0.00       | 0.39 ± 0.00      | 0.24 ± 0.00      | 0.02 ± 0.00      | 0.11 ± 0.00      | 0.20 ± 0.13     |
| DAMP     | P  | –                 | –                | –                | 0.35 ± 0.41      | –                |                 |
|          | R  | –                 | –                | –                | 0.51 ± 0.50      | –                |                 |
|          | F1 | –                 | –                | –                | **0.40 ± 0.43**  | –                | –               |

are recent. In particular, they have been introduced between 2018 and early 2022. (3) Impact: the chosen algorithms have been published in top-tier conferences and are highly cited papers from the field. (4) Code availability: the official codes of the selected algorithms are publicly available. (5) Diversity: methods from different paradigms have been considered. The only paradigm that was ignored is the distance-based since we were not able to find a deep learning approach falling in this category.

## 5. Results

### 5.1. Performance using standard metrics

**Best performing approach.** Table 4 reports the performance of the evaluated methods using standard metrics (precision, recall, and F1-score) on the five considered datasets. The performance is first averaged over the data of every dataset, which in turn is averaged over 5 runs.

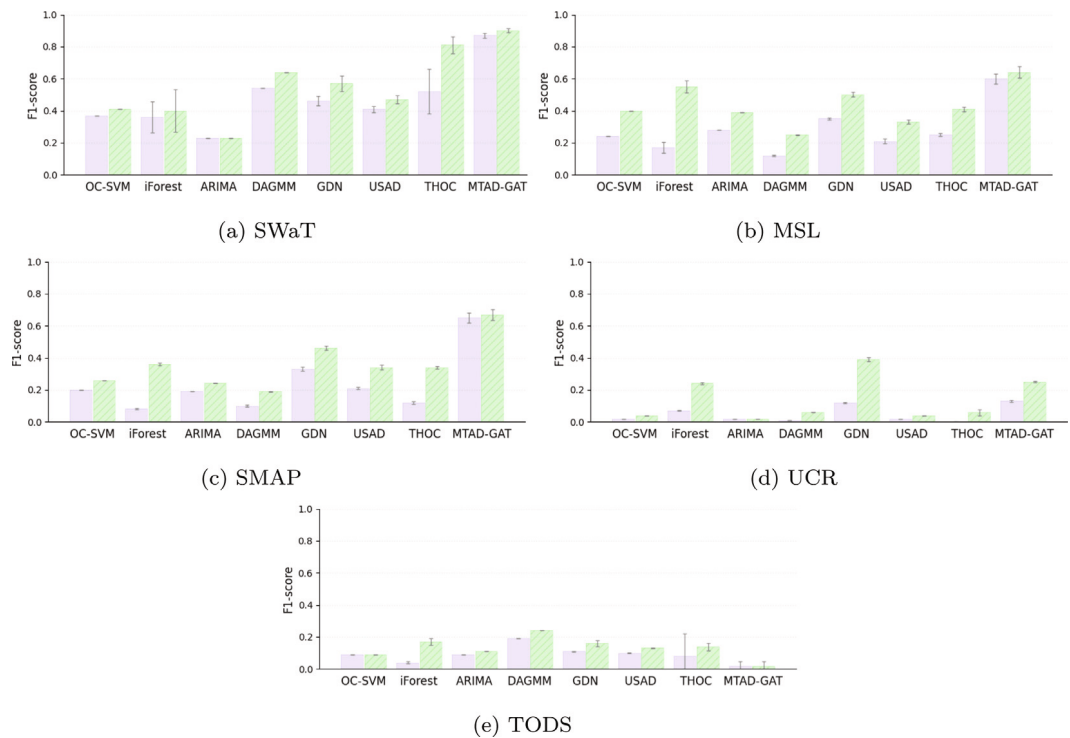

**Fig. 6.** Mean F1-Score on the five datasets. The non-hatched and hatched bars correspond to the mean F1-Score with and without Point Adjustment (PA), respectively. The vertical black line represents the standard deviation over five runs.

For a more intuitive visualization of the results, Fig. 6 shows the F1-score with and without Point Adjustment (PA). In general, MTAD-GAT (Zhao et al., 2020) is the best-performing approach, surpassing other methods on three datasets, namely, SWaT, MSL, SMAP. This can be explained by the fact that this approach is hybrid as it is based on forecasting and reconstruction losses. Indeed, this allows the simultaneous detection of local and global anomalies. Nevertheless, it can be remarked that the results obtained on TODS contradict this statement. Indeed, MTAD-GAT registers inferior performance on this benchmark as compared to other methods, including DL and conventional methods. Two hypotheses might justify this drop: (*i*) the TODS dataset encloses complex anomalies that are *moderately local* and are hardly captured by a simple forecasting and/or reconstruction approach, favoring probabilistic modeling as in DAGMM (Zong et al., 2018). However, the higher performance obtained for GDN (Deng & Hooi, 2021) and USAD (Audibert et al., 2020) partly disprove this assumption; and (*ii*) the synthetic data in TODS are not realistic, making them hardly predictable. Another observation that can be made is that GDN (Deng & Hooi, 2021) presents the second-best performance on two datasets, namely MSL, SMAP. This confirms the relevance of using graph representations for modeling time-series. Surprisingly, graph-based approaches (GDN and MTAD-GAT) remain relatively effective on a univariate dataset (UCR), although modeling the connectivity between variables is unnecessary.

**DL vs conventional methods.** As reported in Table 4 and Fig. 6, DL methods, specifically MTAD-GAT and GDN, generally outperform conventional methods. For example, the superiority of DL approaches is extremely noticeable when comparing GDN and ARIMA, which are both forecasting techniques. This increase in performance can be explained by the fact that ARIMA struggles to model the dependencies between variables. However, the gap in performance between DL and traditional methods is less visible in some cases, supporting the assumption of Wu and Keogh (2021) and contradicting (Choi et al., 2021), which argues that DL methods are more effective in the presence of high-dimensional time-series. For instance, DAMP (Lu et al., 2022) beats all DL methods by a large margin showing an average F1-score of 0.28 against only 0.13 for the best-performing DL-technique. OC-SVM shows comparable performance with several DL-based anomaly methods such as THOC and USAD on high-dimensional datasets. More precisely, OC-SVM achieves an F1-score of 0.24 and 0.2 on MSL and MSAP against 0.21 and 0.21 for USAD and 0.25 and 0.2 for THOC, respectively. Another observation that can be made is that conventional approaches, except iForest, seem to be suitable for applications where recall is more important than precision. An example of such an application could be the detection of debris among other objects in space (Musallam et al., 2021). On the contrary, DL approaches are overall more precise.

**Impact of point adjustment.** From the results of Table 5, and Fig. 6, it can be noted that the Point Adjustment (PA) process significantly boosts the performance. In particular, the highest performance gain can be observed for iForest on the MSL dataset, where the F1-score increases from 17% to 55%. This can be explained by the fact that PA adjusts the predictions before computing the metrics. The adjustment is made in a way that rewards a detector when detecting at least one instance of an anomalous segment. The intuition behind that is that finding one anomaly in a segment is sufficient for a timely reaction. Such an intuition closely impacts the recall since it increases the number of False Positives (FP). However, as discussed in Kim et al. (2022), using PA can induce a misleading ranking of the performance. This is confirmed in Table 4 where the results obtained for USAD, THOC and DAGMM are comparable and contradict the performance metrics reported in Table 5. In addition, before applying PA, all DL approaches seem to be in general, more effective than conventional approaches. However, after PA, this is no longer the case. For example, iForest achieves comparable performance with THOC. Overall, PA seems to bias the analysis as it treats range-based data as punctual, neglecting the overlap size and the location of anomalies. Consequently, this blurs the applicability of detectors in real-life setups.

**Benchmark complexity.** All tested methods fail to detect effectively anomalies in UCR, although it is a univariate dataset. This might be due to its low ratio of anomalies. As discussed in Section 4.1, UCR is among the first benchmark to mimic a more realistic configuration, highlighting the difficulty of detecting rare anomalies. The rate of anomalies might have a significant role in defining the complexity of a given dataset. For natural anomalies, two observations could be made. First, the average performance on MSL and SMAP is comparable despite having a significantly different number of variables. Second, natural but induced/forced anomalies seem easy to detect, given that all methods perform well on the SWaT dataset. Unfortunately, such a scenario is unrealistic in most real-world settings as the anomalies are generally infrequent. This point was also raised by Wu and Keogh (2021) highlighting that several benchmark datasets have unrealistic anomaly densities.

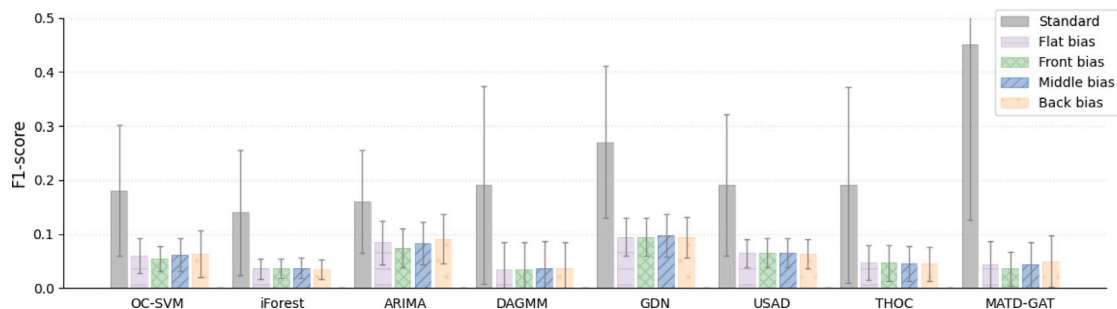

**Fig. 7.** The mean performance per method on all datasets using the range-based metrics of Tatbul et al. (2018), with different location biases.

### 5.2. Performance using revisited metrics for time-series

**Conventional metrics vs revisited metrics.** Table 7 shows the results of evaluation using the revisited F1-score for time-series calculated using $Recall_T$ and $Precision_T$ proposed by Tatbul et al. (2018).

It can be remarked that there exists a significant gap in performance between the results based on conventional and revisited metrics. One main reason is that the revisited metrics consider the overlap size between the predicted sequences and the ground-truth. In contrast, the traditional metrics do not take into account the sequential aspect nor quantify the overlap between the predictions and the ground-truth. Moreover, it can be noted that the results of the revisited metrics are not in full accordance with the conventional ones except for DAMP on UCR. On the one hand, GDN and USAD achieve more competitive results as compared to other approaches. On the other hand, the assumption about the superiority of the hybrid method no longer holds. Overall, both classical and DL forecasting-based techniques give the highest performance. This perspective suggests that the majority of anomalies in benchmarked datasets are local. Finally, DAGMM seems to be among the least effective methods, suggesting its inability to model the distribution of anomalies. This can be explained by the fact that anomalies do not necessarily follow a multimodal Gaussian distribution. Some observations can be made regarding the difficulty of each dataset. First, MSL seems slightly less challenging than SWaT. Second, in line with the results based on conventional metrics, the obtained performance suggests that natural anomalies are more straightforward to detect than synthetic ones. Two reasons could potentially explain that: (1) the datasets with natural anomalies have a high percentage of anomalies, and (2) synthetic datasets do not reliably reflect reality and do not include a sufficient number of anomalies.

**Location bias.** Herein, we analyze the results for different location biases. Table 7, Table 8, Table 9, and Table 10 show the results using flat, front, middle and back bias, respectively. As mentioned in Section 4.2, taking into account the size, the cardinality and the location of the overlap between a predicted sequence and its corresponding ground truth is crucial. Therefore, the location bias weights every predicted time-stamp given its location in the sequence.

Fig. 7 depicts the overall performance for each method under different bias settings. Two observations can be made: (*i*) Although the idea of location bias seems theoretically interesting and flexible for different domain-specific applications, it does not practically bring more information in our experiments, as the average F1-score does not change importantly. However, the most notable results are registered for the middle and back biases as compared to the flat and front biases. This can be attributed to the uneven distribution of anomalies in datasets like SMAP and MSL. As noted by Wu and Keogh (2021), most anomalies in these datasets occur towards the end of the sequences. This may explain the improved performance of all the evaluated methods on MSL and SMAP, particularly for middle and back location biases. This also suggests that most detectors are less mature for applications necessitating an early anomaly detection such as real-time intrusion detection (Zhang, Cushing, de Laat, & Grosso, 2021), cyberattack attempts via network activity (Siddiqui et al., 2019) or cancer detection (Kourou et al., 2015). (*ii*) Additionally, all range-based metrics results are less impressive than the ones obtained using standard metrics. This drop in performance may suggest that most approaches perform poorly in identifying the overlap size and cardinality between a predicted sequence and its corresponding ground truth. In other words, a predicted sequence does not perfectly align with its corresponding ground truth sequence, as the boundaries of anomaly sequences are not well predicted.

**DL vs conventional methods.** Although the top-three best-performing methods are DL models (according to the conventional metrics), it can be seen that classical approaches such as DAMP can outperform DL-methods by a large margin, with advantage of a stable model. Similarly, OC-SVM can achieve comparable performance with its counterpart clustering DL approach, namely THOC. This suggests that conventional methods are not obsolete and that, depending on the application, they can be considered for anomaly detection (Wu & Keogh, 2021).

**Univariate vs multivariate.** The results of Table 4 and Table 7 seem to be in accordance. Indeed, all methods except DAMP (Lu et al., 2022) seem to have poor performance on UCR, which is univariate, while on other multivariate datasets such as MSL, the performance is relatively higher.

### 5.3. Model stability

Besides reporting the precision, recall, and F1-score, it is interesting to observe the behavior of detectors when trained with different initializations. Table 4 and Fig. 6 report the performance average and standard deviation for every approach after five runs. Undoubtedly the most stable methods are the deterministic ones which are ARIMA, OC-SVM, and DAMP. Among DL approaches, DAGMM seems to be the most stable. This could be explained by the fact that it is a density-based approach that relies on estimating the density of normal data. In contrast, THOC and iForest achieve less stable results, especially on SWaT.

Table 6
Number of parameters and model size in Mega Bytes (MB) of the trained models on the different datasets.

| Model | SWaT | | MSL | | SMAP | | UCR | | TODS | |
|---|---|---|---|---|---|---|---|---|---|---|
| | Parameters | Model size (Mb) | Parameters | Model size (Mb) | Parameters | Model size (Mb) | Parameters | Model size (Mb) | Parameters | Model size (Mb) |
| USAD | 1.256.871 | 4,79 | 1.414.755 | 5,40 | 441.225 | 1,68 | 12.321 | 0,05 | 136.710 | 0,54 |
| GDN | 4.225 | 0,02 | 4.481 | 0,02 | 2.561 | 0,01 | 1.025 | 0,01 | 1.601 | 0,01 |
| THOC | 104.768 | 0,41 | 105.792 | 0,42 | 98.112 | 0,39 | 91.968 | 0,36 | 94.272 | 0,37 |
| MTAD-GAT | 373.637 | 1,62 | 384.145 | 1,66 | 314.695 | 1,39 | 274.687 | 1,05 | 288.070 | 1,05 |
| DAGMM | 266.930 | 1,50 | 270.542 | 1,50 | 243.452 | 1,40 | 221.780 | 1,30 | 288.070 | 1,40 |

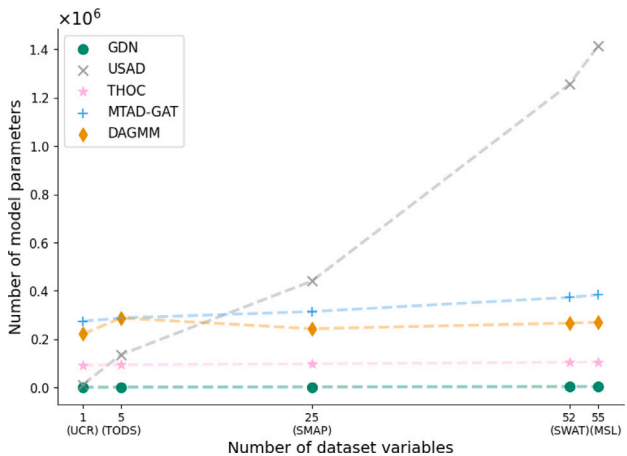

**Fig. 8.** Relation between the number of the parameters of the model and the number of features in the considered dataset.

*5.4. Model size and memory consumption*

Table 6 depicts the number of parameters and the model size in Mega Bytes (MB) of the tested DL architectures. GDN seems to have the lowest number of parameters when tested on all the datasets. In fact, the number of parameters in USAD is around 300 times higher than GDN. This is explained by the fact that the architecture of USAD is complex and is composed of two adversarially trained auto-encoders. In Fig. 8, it can also be seen that contrary to other models, which vary almost linearly, the number of parameters increases at a considerably higher rate. Additionally, despite the significant difference in parameter number, GDN still achieves better results than USAD. This highlights the relevance of using graph representations not only for modeling time-series but also for building less complex model architectures.

*5.5. Generalization to different types of anomalies*

Fig. 9 shows the percentage of detected anomalies per type for all the tested methods. In general, it can be noted that for the majority of tested techniques, collective trend anomalies are probably the most challenging to detect. ARIMA and GDN, which are predictive approaches, show the best generalization capacity to different types of anomalies. OC-SVM easily detects global point and collective shapelet anomalies but still presents decent results for other anomaly types. The results obtained for USAD suggest that it is more robust to collective outliers (e.g., collective shapelet and seasonality), which can be explained by the fact that it is a reconstruction approach that can essentially capture global inconsistencies. DAGMM effectively detects collective seasonal anomalies but shows less impressive results for collective shape outliers. Again, this might return to the probabilistic nature of DAGMM, which is coupled with a sliding window. Finally, MTAD-GAT fails in detecting collective anomalies, despite being hybrid.

*5.6. Discussion*

In the following, we summarize the main findings of the present evaluation study:

(*i*) It is generally difficult to vote for a best-performing approach or paradigm and the performance of an approach highly depends on the considered use case and the nature of the encountered anomalies. For instance, although the hybrid approach MTAD-GAT seems to outperform most other methods, they also exhibit limitations, such as their unsuitability for detecting collective shapelet and trend anomalies. This highlights the need for researchers to explicitly discuss the specific settings or applications under which their algorithms are effective, ensuring that practitioners understand the circumstances in which these methods should be considered (Wu & Keogh, 2021).

(*ii*) The considered forecasting approaches tend to have the most consistent range-based performance with respect to standard metrics.

(*iii*) Traditional approaches are not necessarily obsolete; in some cases, they can achieve performances that are comparable to DL methods. This supports the claim of Wu and Keogh (2021) to question the assumption that deep learning is the definitive solution for time-series anomaly detection. Since they are usually recall-oriented, they usually detect most types of anomalies but at the cost of a higher false positive rate.

(*iv*) The Point Adjustment (PA) protocol is unreliable as it overestimates the detector performance, and in the case of traditional approaches that are already recall-oriented, this triggers an even higher false positive rate.

(*v*) Multivariate time-series are challenging due to the high dimensionality of data. On the other hand, univariate time-series can be challenging when the anomaly ratio is very low.

(*vi*) Most models achieve low performance using range-based metrics, highlighting the difficulty of detecting the anomaly boundaries.

(*vii*) Model stability and memory consumption can vary importantly from one method to another. Hence, depending on the end-goal application, these metrics can be essential for selecting the most suitable model in accordance with the hardware specifications.

**6. Conclusion**

This paper proposes an extensive evaluation study of recent time-series anomaly detection methods. To the best of our knowledge, we are the first to analyze these algorithms based on a more elaborate experimentation protocol. In contrast to previous evaluation studies, which only consider the standard performance metrics, we take into account revisited performance criteria specifically designed for time series in our analysis. In addition, the model stability, the model size as well as

**Table 7**
The *flat-bias* performance of the tested methods on the 5 benchmarks using the metrics proposed by Tatbul et al. (2018). The average and the standard deviation of five runs are reported.

| | | USAD | GDN | THOC | MTAD-GAT | DAGMM | OC-SVM | iForest | ARIMA | DAMP |
|---|---|---|---|---|---|---|---|---|---|---|
| SWaT | $P_T$ | 0.0583 ± 0.0210 | 0.0703 ± 0.0173 | 0.2905 ± 0.1113 | 0.0331 ± 0.0082 | 0.1032 ± 0.0022 | 0.0312 ± 0.0000 | 0.0038 ± 0.0014 | 0.0395 ± 0.0000 | – |
| | $R_T$ | 0.3284 ± 0.0396 | 0.2505 ± 0.0290 | 0.0204 ± 0.0108 | 0.3946 ± 0.0385 | 0.1889 ± 0.0022 | 0.8566 ± 0.0000 | 0.9021 ± 0.0014 | 0.8085 ± 0.0000 | – |
| | F1 | 0.0976 ± 0.0296 | 0.1081 ± 0.0188 | 0.0381 ± 0.0196 | 0.0607 ± 0.0133 | 0.1334 ± 0.0026 | 0.0602 ± 0.0000 | 0.0077 ± 0.0029 | 0.0753 ± 0.0000 | – |
| MSL | $P_T$ | 0.1210 ± 0.0079 | 0.2494 ± 0.0174 | 0.1592 ± 0.0143 | 0.0768 ± 0.0045 | 0.1109 ± 0.0025 | 0.0528 ± 0.0000 | 0.0389 ± 0.0123 | 0.0843 ± 0.0000 | – |
| | $R_T$ | 0.2014 ± 0.0263 | 0.2293 ± 0.0217 | 0.2234 ± 0.0213 | 0.1846 ± 0.0045 | 0.0379 ± 0.0025 | 0.8316 ± 0.0000 | 0.5846 ± 0.0123 | 0.4106 ± 0.0000 | – |
| | F1 | 0.0666 ± 0.0030 | 0.1535 ± 0.0133 | 0.0770 ± 0.0044 | 0.0319 ± 0.0023 | 0.0161 ± 0.0004 | 0.0856 ± 0.0000 | 0.0625 ± 0.0138 | 0.0937 ± 0.0000 | – |
| SMAP | $P_T$ | 0.1204 ± 0.0095 | 0.1500 ± 0.0032 | 0.0814 ± 0.0047 | 0.1401 ± 0.0000 | 0.0709 ± 0.0094 | 0.0656 ± 0.0000 | 0.0178 ± 0.0064 | 0.1401 ± 0.0000 | – |
| | $R_T$ | 0.2501 ± 0.0070 | 0.2023 ± 0.0160 | 0.0807 ± 0.0081 | 0.5272 ± 0.0000 | 0.0379 ± 0.0094 | 0.8365 ± 0.0000 | 0.3728 ± 0.0064 | 0.5272 ± 0.0000 | – |
| | F1 | 0.0634 ± 0.0032 | 0.0793 ± 0.0033 | 0.0259 ± 0.0018 | 0.1179 ± 0.0000 | 0.0091 ± 0.0009 | 0.1040 ± 0.0000 | 0.0285 ± 0.0064 | 0.1179 ± 0.0000 | – |
| UCR | $P_T$ | 0.0106 ± 0.0002 | 0.0749 ± 0.0032 | 0.0106 ± 0.0040 | 0.0113 ± 0.0017 | 0.0709 ± 0.0094 | 0.0083 ± 0.0000 | 0.0255 ± 0.0034 | 0.0074 ± 0.0000 | 0.3249 ± 0.0000 |
| | $R_T$ | 0.2320 ± 0.0023 | 0.2501 ± 0.0118 | 0.0011 ± 0.0040 | 0.0189 ± 0.0017 | 0.0379 ± 0.0094 | 0.8356 ± 0.0000 | 0.4089 ± 0.0034 | 0.2907 ± 0.0000 | 0.3420 ± 0.0000 |
| | F1 | 0.0168 ± 0.0003 | 0.0473 ± 0.0013 | 0.0016 ± 0.0005 | 0.0054 ± 0.0005 | 0.0091 ± 0.0009 | 0.0134 ± 0.0000 | 0.0352 ± 0.0031 | 0.0119 ± 0.0000 | 0.2847 ± 0.0000 |
| TODS | $P_T$ | 0.0420 ± 0.0039 | 0.0537 ± 0.0130 | 0.0617 ± 0.0047 | 0.0806 ± 0.0970 | 0.0709 ± 0.0094 | 0.0197 ± 0.0000 | 0.0975 ± 0.0862 | 0.0685 ± 0.0000 | – |
| | $R_T$ | 0.4796 ± 0.0015 | 0.5546 ± 0.1713 | 0.1766 ± 0.0047 | 0.0019 ± 0.0970 | 0.0379 ± 0.0094 | 0.8054 ± 0.0000 | 0.0645 ± 0.0862 | 0.6395 ± 0.0000 | – |
| | F1 | 0.0767 ± 0.0065 | 0.0843 ± 0.0147 | 0.0887 ± 0.0073 | 0.0027 ± 0.0035 | 0.0091 ± 0.0009 | 0.0382 ± 0.0000 | 0.0435 ± 0.0041 | 0.1233 ± 0.0000 | – |
| Avg. F1 | | 0.0642 ± 0.0266 | 0.0945 ± 0.0353 | 0.0463 ± 0.0323 | 0.0437 ± 0.0426 | 0.0354 ± 0.0491 | 0.0603 ± 0.0324 | 0.0355 ± 0.0180 | 0.0844 ± 0.0402 | – |

**Table 8**
The *front-bias* performance of the tested methods on the 5 benchmarks using the metrics proposed by Tatbul et al. (2018). The average and the standard deviation of five runs are reported.

|  |  | USAD | GDN | THOC | MTAD-GAT | DAGMM | OC-SVM | iForest | ARIMA | DAMP |
|---|---|---|---|---|---|---|---|---|---|---|
| SWaT | $P_T$ | 0.0590 ± 0.0213 | 0.0705 ± 0.0173 | 0.2905 ± 0.1114 | 0.0312 ± 0.0076 | 0.1046 ± 0.0023 | 0.0331 ± 0.0000 | 0.0039 ± 0.0014 | 0.0404 ± 0.0000 | – |
|  | $R_T$ | 0.3201 ± 0.0348 | 0.2504 ± 0.0296 | 0.0221 ± 0.0113 | 0.4632 ± 0.0480 | 0.1816 ± 0.0023 | 0.8571 ± 0.0000 | 0.9028 ± 0.0014 | 0.8119 ± 0.0000 | – |
|  | F1 | 0.0981 ± 0.0296 | 0.1084 ± 0.0189 | 0.0409 ± 0.0204 | 0.0580 ± 0.0129 | 0.1327 ± 0.0026 | 0.0638 ± 0.0000 | 0.0079 ± 0.0029 | 0.0771 ± 0.0000 | – |
| MSL | $P_T$ | 0.1210 ± 0.0081 | 0.2494 ± 0.0176 | 0.1631 ± 0.0150 | 0.0758 ± 0.0046 | 0.1110 ± 0.0025 | 0.0527 ± 0.0000 | 0.0407 ± 0.0132 | 0.0690 ± 0.0000 | – |
|  | $R_T$ | 0.2114 ± 0.0262 | 0.2293 ± 0.0217 | 0.1996 ± 0.0192 | 0.1871 ± 0.0046 | 0.0291 ± 0.0025 | 0.8125 ± 0.0000 | 0.5542 ± 0.0132 | 0.4230 ± 0.0000 | – |
|  | F1 | 0.0671 ± 0.0023 | 0.1535 ± 0.0133 | 0.0806 ± 0.0047 | 0.0308 ± 0.0029 | 0.0156 ± 0.0005 | 0.0813 ± 0.0000 | 0.0627 ± 0.0140 | 0.0727 ± 0.0000 | – |
| SMAP | $P_T$ | 0.1202 ± 0.0095 | 0.1509 ± 0.0031 | 0.0817 ± 0.0048 | 0.1102 ± 0.0000 | 0.0708 ± 0.0094 | 0.0363 ± 0.0000 | 0.0192 ± 0.0071 | 0.1102 ± 0.0000 | – |
|  | $R_T$ | 0.2634 ± 0.0079 | 0.1913 ± 0.0164 | 0.0690 ± 0.0123 | 0.5410 ± 0.0000 | 0.0388 ± 0.0094 | 0.8238 ± 0.0000 | 0.3529 ± 0.0071 | 0.5410 ± 0.0000 | – |
|  | F1 | 0.0642 ± 0.0025 | 0.0761 ± 0.0030 | 0.0227 ± 0.0013 | 0.0843 ± 0.0000 | 0.0090 ± 0.0011 | 0.0648 ± 0.0000 | 0.0296 ± 0.0061 | 0.0843 ± 0.0000 | – |
| UCR | $P_T$ | 0.0106 ± 0.0002 | 0.0753 ± 0.0030 | 0.0106 ± 0.0040 | 0.0113 ± 0.0017 | 0.0708 ± 0.0094 | 0.0079 ± 0.0000 | 0.0270 ± 0.0035 | 0.0076 ± 0.0000 | 0.3312 ± 0.0000 |
|  | $R_T$ | 0.2295 ± 0.0029 | 0.2510 ± 0.0109 | 0.0011 ± 0.0040 | 0.0190 ± 0.0017 | 0.0388 ± 0.0094 | 0.8286 ± 0.0000 | 0.3895 ± 0.0035 | 0.2911 ± 0.0000 | 0.3330 ± 0.0000 |
|  | F1 | 0.0167 ± 0.0002 | 0.0496 ± 0.0010 | 0.0015 ± 0.0005 | 0.0051 ± 0.0003 | 0.0090 ± 0.0011 | 0.0145 ± 0.0000 | 0.0362 ± 0.0032 | 0.0121 ± 0.0000 | 0.2588 ± 0.0000 |
| TODS | $P_T$ | 0.0418 ± 0.0042 | 0.0532 ± 0.0122 | 0.0620 ± 0.0042 | 0.0801 ± 0.097 | 0.0708 ± 0.0094 | 0.0247 ± 0.0000 | 0.0979 ± 0.0865 | 0.0679 ± 0.0000 | – |
|  | $R_T$ | 0.4795 ± 0.0115 | 0.5549 ± 0.1724 | 0.1775 ± 0.0042 | 0.0019 ± 0.0970 | 0.0388 ± 0.0094 | 0.8028 ± 0.0000 | 0.0656 ± 0.0865 | 0.6396 ± 0.0000 | – |
|  | F1 | 0.0767 ± 0.0065 | 0.0837 ± 0.0136 | 0.0886 ± 0.0075 | 0.0023 ± 0.0035 | 0.0090 ± 0.0011 | 0.0472 ± 0.0000 | 0.0437 ± 0.0035 | 0.1222 ± 0.0000 | – |
| Avg. F1 |  | 0.0646 ± 0.0267 | 0.0943 ± 0.0351 | 0.0469 ± 0.0333 | 0.0361 ± 0.0314 | 0.0351 ± 0.0489 | 0.0543 ± 0.0226 | 0.0360 ± 0.0179 | 0.0737 ± 0.0354 | – |

**Table 9**
The *middle-bias* performance of the tested methods on the 5 benchmarks using the metrics proposed by Tatbul et al. (2018). The average and the standard deviation of five runs are reported.

| | | USAD | GDN | THOC | MTAD-GAT | DAGMM | OC-SVM | iForest | ARIMA | DAMP |
|---|---|---|---|---|---|---|---|---|---|---|
| SWaT | $P_T$ | 0.0592 ± 0.0214 | 0.0705 ± 0.0173 | 0.2905 ± 0.1113 | 0.0344 ± 0.0085 | 0.1040 ± 0.0022 | 0.0337 ± 0.0000 | 0.0040 ± 0.0015 | 0.0382 ± 0.0000 | – |
| | $R_T$ | 0.3382 ± 0.0432 | 0.2523 ± 0.0301 | 0.0217 ± 0.0131 | 0.4391 ± 0.0431 | 0.2018 ± 0.0022 | 0.8571 ± 0.0000 | 0.9028 ± 0.0015 | 0.8057 ± 0.0000 | – |
| | F1 | 0.0993 ± 0.0303 | 0.1084 ± 0.0188 | 0.0401 ± 0.0236 | 0.0634 ± 0.0140 | 0.1372 ± 0.0025 | 0.0648 ± 0.0000 | 0.0080 ± 0.0030 | 0.0731 ± 0.0000 | – |
| MSL | $P_T$ | 0.1208 ± 0.0081 | 0.2518 ± 0.0180 | 0.1590 ± 0.0141 | 0.0760 ± 0.0046 | 0.1109 ± 0.0025 | 0.0567 ± 0.0000 | 0.0410 ± 0.0133 | 0.0831 ± 0.0000 | – |
| | $R_T$ | 0.2221 ± 0.0330 | 0.2547 ± 0.0222 | 0.2379 ± 0.0248 | 0.1846 ± 0.0046 | 0.0405 ± 0.0025 | 0.8491 ± 0.0000 | 0.6078 ± 0.0133 | 0.4078 ± 0.0000 | – |
| | F1 | 0.0666 ± 0.0037 | 0.1652 ± 0.0176 | 0.0741 ± 0.0043 | 0.0301 ± 0.0023 | 0.0152 ± 0.0004 | 0.0927 ± 0.0000 | 0.0655 ± 0.0149 | 0.0891 ± 0.0000 | – |
| SMAP | $P_T$ | 0.1204 ± 0.0096 | 0.1506 ± 0.0032 | 0.0800 ± 0.0050 | 0.1363 ± 0.0000 | 0.0708 ± 0.0094 | 0.0582 ± 0.0000 | 0.0189 ± 0.0070 | 0.1363 ± 0.0000 | – |
| | $R_T$ | 0.2608 ± 0.0096 | 0.2254 ± 0.0179 | 0.0813 ± 0.0047 | 0.5403 ± 0.0000 | 0.0416 ± 0.0094 | 0.8484 ± 0.0000 | 0.3890 ± 0.0070 | 0.5403 ± 0.0000 | – |
| | F1 | 0.0656 ± 0.0039 | 0.0847 ± 0.0025 | 0.0238 ± 0.0018 | 0.1146 ± 0.0000 | 0.0089 ± 0.0010 | 0.0962 ± 0.0000 | 0.0301 ± 0.0069 | 0.1146 ± 0.0000 | – |
| UCR | $P_T$ | 0.0104 ± 0.0002 | 0.0759 ± 0.0029 | 0.0106 ± 0.0040 | 0.0114 ± 0.0017 | 0.0708 ± 0.0094 | 0.0090 ± 0.0000 | 0.0269 ± 0.0035 | 0.0075 ± 0.0000 | 0.3325 ± 0.0000 |
| | $R_T$ | 0.2375 ± 0.0024 | 0.2524 ± 0.0123 | 0.0015 ± 0.0040 | 0.0201 ± 0.0017 | 0.0416 ± 0.0094 | 0.8421 ± 0.0000 | 0.4258 ± 0.0035 | 0.2920 ± 0.0000 | 0.3607 ± 0.0000 |
| | F1 | 0.0165 ± 0.0003 | 0.0439 ± 0.0019 | 0.0019 ± 0.0008 | 0.0054 ± 0.0006 | 0.0089 ± 0.0010 | 0.0151 ± 0.0000 | 0.0371 ± 0.0032 | 0.0121 ± 0.0000 | 0.2940 ± 0.0000 |
| TODS | $P_T$ | 0.0426 ± 0.0040 | 0.0541 ± 0.0131 | 0.0622 ± 0.0046 | 0.0802 ± 0.0970 | 0.0708 ± 0.0094 | 0.0204 ± 0.0000 | 0.0972 ± 0.0862 | 0.0682 ± 0.0000 | – |
| | $R_T$ | 0.4791 ± 0.0116 | 0.5557 ± 0.1702 | 0.1772 ± 0.0046 | 0.0018 ± 0.0970 | 0.0416 ± 0.0094 | 0.8065 ± 0.0000 | 0.0645 ± 0.0862 | 0.6405 ± 0.0000 | – |
| | F1 | 0.0777 ± 0.0066 | 0.0847 ± 0.0144 | 0.0894 ± 0.0073 | 0.0023 ± 0.0035 | 0.0089 ± 0.0010 | 0.0397 ± 0.0000 | 0.0434 ± 0.0039 | 0.1227 ± 0.0000 | – |
| Avg. F1 | | 0.0651 ± 0.0272 | 0.0976 ± 0.0397 | 0.0459 ± 0.0321 | 0.0432 ± 0.0419 | 0.0358 ± 0.0507 | 0.0617 ± 0.0310 | 0.0368 ± 0.0187 | 0.0823 ± 0.0393 | – |

**Table 10**
The *back-bias* performance of the tested methods on the 5 benchmarks using the metrics proposed by Tatbul et al. (2018). The average and the standard deviation of five runs are reported.

| | | USAD | GDN | THOC | MTAD-GAT | DAGMM | OC-SVM | iForest | ARIMA | DAMP |
|---|---|---|---|---|---|---|---|---|---|---|
| SWaT | $P_T$ | 0.0576 ± 0.0208 | 0.0701 ± 0.0174 | 0.2905 ± 0.1113 | 0.0351 ± 0.0087 | 0.1018 ± 0.0022 | 0.0292 ± 0.0000 | 0.0037 ± 0.0014 | 0.0385 ± 0.0000 | – |
| | $R_T$ | 0.3367 ± 0.0449 | 0.2505 ± 0.0286 | 0.0188 ± 0.0103 | 0.3261 ± 0.0291 | 0.1961 ± 0.0022 | 0.8561 ± 0.0000 | 0.9014 ± 0.0014 | 0.8050 ± 0.0000 | – |
| | F1 | 0.0969 ± 0.0296 | 0.1078 ± 0.0187 | 0.0352 ± 0.0189 | 0.0628 ± 0.0135 | 0.1340 ± 0.0026 | 0.0565 ± 0.0000 | 0.0075 ± 0.0029 | 0.0734 ± 0.0000 | – |
| MSL | $P_T$ | 0.1231 ± 0.0073 | 0.2494 ± 0.0172 | 0.1553 ± 0.0137 | 0.0778 ± 0.0044 | 0.1107 ± 0.0025 | 0.0528 ± 0.0000 | 0.0372 ± 0.0114 | 0.0996 ± 0.0000 | – |
| | $R_T$ | 0.1914 ± 0.0264 | 0.2405 ± 0.0201 | 0.2471 ± 0.0245 | 0.1820 ± 0.0044 | 0.0467 ± 0.0025 | 0.8507 ± 0.0000 | 0.6151 ± 0.0114 | 0.3982 ± 0.0000 | – |
| | F1 | 0.0648 ± 0.0037 | 0.1558 ± 0.0130 | 0.0706 ± 0.0037 | 0.0325 ± 0.0016 | 0.0164 ± 0.0164 | 0.0887 ± 0.0000 | 0.0613 ± 0.0137 | 0.1076 ± 0.0000 | – |
| SMAP | $P_T$ | 0.1206 ± 0.0025 | 0.1491 ± 0.0033 | 0.0811 ± 0.0047 | 0.1699 ± 0.0000 | 0.0711 ± 0.0094 | 0.0949 ± 0.0000 | 0.0164 ± 0.0057 | 0.1699 ± 0.0000 | – |
| | $R_T$ | 0.2367 ± 0.0070 | 0.2133 ± 0.0157 | 0.0923 ± 0.0042 | 0.5133 ± 0.0000 | 0.0369 ± 0.0094 | 0.8491 ± 0.0000 | 0.3927 ± 0.0057 | 0.5133 ± 0.0000 | – |
| | F1 | 0.0612 ± 0.0025 | 0.0807 ± 0.0033 | 0.0275 ± 0.0022 | 0.1375 ± 0.0000 | 0.0092 ± 0.0008 | 0.1330 ± 0.0000 | 0.0270 ± 0.0065 | 0.1375 ± 0.0000 | – |
| UCR | $P_T$ | 0.0106 ± 0.0002 | 0.0744 ± 0.0035 | 0.0106 ± 0.0040 | 0.0114 ± 0.0017 | 0.0711 ± 0.0094 | 0.0087 ± 0.0000 | 0.0239 ± 0.0032 | 0.0073 ± 0.0000 | 0.3186 ± 0.0000 |
| | $R_T$ | 0.2345 ± 0.0020 | 0.2492 ± 0.0126 | 0.0011 ± 0.0040 | 0.0188 ± 0.0017 | 0.0369 ± 0.0094 | 0.8426 ± 0.0000 | 0.4283 ± 0.0032 | 0.2903 ± 0.0000 | 0.3509 ± 0.0000 |
| | F1 | 0.0167 ± 0.0003 | 0.0411 ± 0.0018 | 0.0017 ± 0.0006 | 0.0054 ± 0.0006 | 0.0092 ± 0.0008 | 0.0121 ± 0.0000 | 0.0337 ± 0.0031 | 0.0116 ± 0.0000 | 0.2668 ± 0.0000 |
| TODS | $P_T$ | 0.0421 ± 0.0039 | 0.0542 ± 0.0140 | 0.0614 ± 0.0063 | 0.0812 ± 0.0970 | 0.0711 ± 0.0094 | 0.0147 ± 0.0000 | 0.0970 ± 0.0859 | 0.0692 ± 0.0000 | – |
| | $R_T$ | 0.4796 ± 0.0116 | 0.5543 ± 0.1702 | 0.1757 ± 0.0063 | 0.0018 ± 0.0970 | 0.0369 ± 0.0094 | 0.8080 ± 0.0000 | 0.0634 ± 0.0859 | 0.6393 ± 0.0000 | – |
| | F1 | 0.0769 ± 0.0065 | 0.0848 ± 0.0158 | 0.0882 ± 0.0084 | 0.0029 ± 0.0037 | 0.0092 ± 0.0008 | 0.0289 ± 0.0000 | 0.0432 ± 0.0046 | 0.1243 ± 0.0000 | – |
| Avg. F1 | | 0.0633 ± 0.0264 | 0.0940 ± 0.0376 | 0.0446 ± 0.0310 | 0.0482 ± 0.0496 | 0.0356 ± 0.0493 | 0.0638 ± 0.0433 | 0.0345 ± 0.0178 | 0.0909 ± 0.0451 | – |

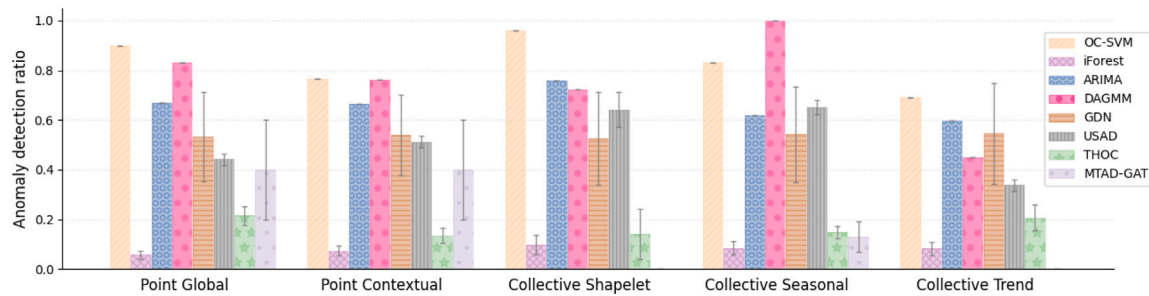

**Fig. 9.** The ratio of true anomalies detected for each tested method when varying the anomaly types. All methods succeeded in partially detecting each anomaly type, except MTAD-GAT which was unable to detect any collective trend anomaly.

the robustness to different types of anomalies are also investigated. All these additional elements give a more complete picture of the current state-of-the-art. Moreover, the proposed protocol is timely and could be beneficial for future investigations, providing more insights regarding their applicability in a real-world context.

**CRediT authorship contribution statement**

**Nesryne Mejri:** Methodology, Data collection, Data analysis, Writing – original draft. **Laura Lopez-Fuentes:** Methodology, Data collection, Data analysis, Writing – original draft. **Kankana Roy:** Methodology. **Pavel Chernakov:** Methodology, Data collection. **Enjie Ghorbel:** Conceptualization, Methodology, Data collection, Data analysis, Writing – original draft, Writing – review & editing, Supervision. **Djamila Aouada:** Writing – review & editing, Funding acquisition, Supervision.

**Declaration of competing interest**

The authors declare the following financial interests/personal relationships which may be considered as potential competing interests: Nesryne Mejri reports financial support was provided by National Research Fund. Laura Lopez-Fuentes reports financial support was provided by National Research Fund. Kankana Roy reports financial support was provided by National Research Fund. Pavel Chernakov reports financial support was provided by National Research Fund. Enjie Ghorbel reports financial support was provided by National Research Fund. Djamila Aouada reports financial support was provided by National Research Fund. Nesryne Mejri reports financial support was provided by Post Luxembourg. Laura Lopez-Fuentes reports financial support was provided by Post Luxembourg. Kankana Roy reports financial support was provided by Post Luxembourg. Pavel Chernakov reports financial support was provided by Post Luxembourg. Enjie Ghorbel reports financial support was provided by Post Luxembourg. Djamila Aouada reports financial support was provided by Post Luxembourg. Laura Lopez-Fuentes reports financial support was provided by European Space Agency. Kankana Roy reports financial support was provided by European Space Agency. Pavel Chernakov reports financial support was provided by European Space Agency. Enjie Ghorbel reports financial support was provided by European Space Agency. Djamila Aouada reports financial support was provided by European Space Agency. Enjie Ghorbel reports a relationship with University of Manouba, Cristal Laboratory, National School of Computer Science that includes: employment. Kankana Roy reports a relationship with Karolinska Institute that includes: employment. Laura Lopez-Fuentes reports a relationship with TotIA that includes: employment. Co-author Laura Lopez-Fuentes is currently working as an engineer at TotIA. Co-author Kankana Roy currently works for the Karolinska Institutet. Co-author Enjie Ghorbel is currently primarily affiliated with University of Manouba, Cristal Laboratory, National School of Computer Science

**Data availability**

No data was used for the research described in the article.

**Acknowledgments**

This work is supported by the Luxembourg National Research Fund (FNR) under the projects BRIDGES2021/IS/16353350/FaKeDeTeR, and UNFAKE, ref. 16763798 both in collaboration with POST Luxembourg and by the European Space Agency (ESA) under the project SKYTRUST 4000133885/21/ NL/MH/hm. Further acknowledgments go to the dataset donors for their efforts in collecting, cleaning, and curating the data.